%% file: main.tex
\documentclass{article}
\usepackage{iclr2027_conference,times}
\iclrfinalcopy 
\input{math_commands.tex}

\usepackage{hyperref}
\usepackage{url}
\usepackage{booktabs}
\usepackage{array}
\usepackage{graphicx}
\usepackage{amsmath,amssymb}
\usepackage{placeins}

\title{Staying on the Attractor:\\ Supervising Neural Surrogates of\\ 3D Turbulence Where They Leave It}

\author{Yilong Dai$^{1}$, Shaswata Mitra$^{1}$, Raj Patel$^{1}$, Yiming Sun$^{2}$, Shengyu Chen$^{3}$, \\
\bfseries Jiaqi Gong$^{1}$, Sudip Mittal$^{1}$, Shahram Rahimi$^{1}$, Xiaowei Jia$^{2}$, Runlong Yu$^{1}$\thanks{Corresponding author: \texttt{ryu5@ua.edu}. Code: \url{https://anonymous.4open.science/r/oas-turbulence-5B16}.} \\[4pt]
\normalfont $^{1}$The University of Alabama \quad $^{2}$Rutgers University \quad $^{3}$University of Pittsburgh}

\begin{document}
\maketitle
\lhead{Preprint}

\begin{abstract}
Neural surrogates are trained to predict 3D turbulent flows in place of direct numerical simulation (DNS). For chaotic flows, the goal is short-term pointwise accuracy followed by long-term physical and statistical fidelity. However, small prediction errors can carry a surrogate away from the flow's \emph{attractor}. Off-attractor states are poorly represented in training data, leaving their evolution weakly constrained. The learned dynamics can then amplify deviations and lead to blow-up, freezing, or statistical drift. In this paper, we propose \emph{off-attractor supervision} (OAS) to supervise neural surrogates where they leave the attractor. OAS teaches the model how the true Navier--Stokes dynamics would evolve from these states. Each selected state is paired with its own future computed by DNS. Three generators select a few hundred states for relabeling. The first collects states from the surrogate's own rollouts. The second uses surrogate attacks to target freezing, excessive amplification, and violations of incompressibility and energy balance. The third perturbs training states along an amplified direction and a strongly damped random direction of the dynamics. All attacks run on the surrogate alone, and DNS relabeling is performed offline once per selected state. Experiments on $128^{3}$ turbulence show that OAS increases the median time to failure from 21 to 721 steps. The compared baselines achieve medians of at most 110 steps, and the advantage holds across training seeds. OAS also achieves the lowest pointwise error at step 15 and the best long-horizon statistics among the compared methods. OAS integrates physical models into neural simulation by extending supervision from fixed reference trajectories to states where the surrogate is likely to fail. This principle can guide the development of more reliable scientific surrogates when deployment takes models beyond the coverage of their training data.
\end{abstract}

\input{sections/1_intro}
\input{sections/2_problem}
\input{sections/3_method}
\input{sections/4_experiments}
\input{sections/5_limitations}
\input{sections/6_conclusion}

\bibliography{references,references_arxiv}
\bibliographystyle{iclr2027_conference}

\appendix
\input{sections/appendix}

\end{document}

%% file: math_commands.tex
\usepackage{amsmath,amsfonts,bm}

\def\eqref#1{equation~\ref{#1}}

\def\1{\bm{1}}

\DeclareMathAlphabet{\mathsfit}{\encodingdefault}{\sfdefault}{m}{sl}
\SetMathAlphabet{\mathsfit}{bold}{\encodingdefault}{\sfdefault}{bx}{n}



%% file: sections/1_intro.tex
\section{Introduction}\label{sec:intro}

Turbulence is a dominant source of simulation cost in combustion, aerodynamics, ocean hydrodynamics, and atmospheric modeling. The three-dimensional incompressible Navier--Stokes equations are nonlinear and chaotic. Unlike their two-dimensional counterparts they stretch vorticity and cascade energy toward small scales, so a simulation must resolve many scales at once. Direct numerical simulation (DNS) resolves every dynamically active scale down to the dissipation scale \citep{pope2000turbulent}. Its cost grows steeply with Reynolds number, since a more turbulent flow demands a finer grid and a shorter time step. Flows of engineering interest lie beyond what a resolved simulation reaches, and neural surrogates are trained to replace it \citep{li2021fno,wu2024transolver,pan2026msmoe,mukhopadhyay2026walrus,armegioiu2026memory,ye2025recurrent,dai2026pestphysicsenhancedswintransformer,dai2026physicspreservinglatentcompressionzeroshot}. The hardest task for a surrogate, and the standard way to test one, is autoregressive rollout. The model is applied to its own output from a single initial state without further ground truth, so its errors compound.

Chaos limits what a long prediction can mean. No rollout tracks the truth beyond a horizon the flow itself sets (Section~\ref{sec:horizons}), so pointwise accuracy is the target only up to that horizon. Beyond it, the prediction should remain a physically admissible turbulent state with the right statistics, the criterion used to evaluate long emulator rollouts in global weather forecasting \citep{lam2023graphcast,price2025gencast,bonev2025fourcastnet3,mahesh2024hens,wattmeyer2024ace2}. This paper addresses both the criterion a long rollout should be evaluated against and the training that has to meet it.

Evaluated that way, every baseline we compared fails, most of them within a turnover time or two (Section~\ref{sec:main}). They are trained on DNS trajectories. Whatever the objective, every state such a trajectory contains lies on the \emph{attractor} of the flow, the set of states the flow itself visits (Section~\ref{sec:setting}). Small errors displace a rollout from that set within a turnover time. The model then receives inputs it has never seen, and the energy blows up, the output freezes, or the statistics drift. The true dynamics does none of this. Over one step it does not amplify a deviation of the kind a surrogate makes, whereas the surrogate amplifies along directions that no clean training pair constrains (Section~\ref{sec:markov}). A mean-squared-error objective also underweights the smallest scales, which degrades the statistics of a rollout without making it fail (Section~\ref{sec:spectra}).

Existing stabilization methods act on the training inputs, on the predicted state, or on the learned operator. The first family trains on deviated inputs, by noise injection \citep{sanchezgonzalez2020learning,stachenfeld2022learned}, the pushforward trick \citep{brandstetter2022message}, unrolled training \citep{um2020solver,kochkov2021ml,list2025differentiability,chakraborty2024multistep}, or sampling toward the states where a model performs poorly \citep{ouyang2025rams}. The second acts on the predicted state, by denoising it \citep{lippe2023pderefiner,kohl2023acdm,dai2026flowrefiner,yoo2026diffusionrollout} or correcting it at inference \citep{pedersen2025thermalizer,liu2026selfrefining}, and the third on the learned operator \citep{mccabe2023stability,pervez2026transient,nie2026jaws,schiff2024dyslim}. Ours also trains on deviated inputs, and the difference lies in \emph{label assignment}. In a dynamical system the target is the future of the input, so it moves when the input does. Noise injection, the pushforward trick, and unrolled training instead keep the future of the clean state as the label. The alternative, pairing a deviated state with the solver's answer at that state, appears in the diverted-chain objective of \citet{koehler2024apebench}, the adversarial loss of \citet{sun2025solver}, and the active-learning scheme of \citet{roy2026beyonduniform}. These approaches query the solver during training or state selection, and some require differentiating through it. Their demonstrations reach $32^{3}$ or two dimensions. We take the solver out of the loop. A non-differentiable DNS is run once per state, offline, after the states are chosen, which leaves the choice of states open and brings three-dimensional turbulence within reach. Appendix~\ref{app:related} places this work against each family.

We therefore propose \emph{off-attractor supervision} (OAS), a data-augmentation method combining failure-mode-guided state selection and offline DNS relabeling for three-dimensional turbulence. A deviated state is paired with its own future, computed by the DNS, so the label is exact wherever the state lies. The surrogate is then supervised where clean data leave it unconstrained, and it learns the response of the dynamics to a deviation instead of producing one of its own. Three generators, which follow the diagnosis above, select the states worth relabeling (Section~\ref{sec:generators}). The first collects the states the surrogate reaches in its own rollouts. The second attacks the surrogate by gradient ascent on the network alone, with one objective per failure mode. The third perturbs training states along a direction the dynamics amplifies and along a random one it strongly damps. The solver labels each state once. The pairs fill a fixed fraction of fine-tuning sample slots.

We make three contributions. \textbf{(C1) Off-attractor supervision.} We pair states off the attractor with their own future under the DNS rather than with the future of the clean state they came from. The solver runs offline, once per state, and is never differentiated through, which brings the construction to $128^{3}$. \textbf{(C2) A diagnosis, and a criterion for failure.} We measure how the dynamics and the learned map each respond to a deviation shaped like a rollout's own error. The amplification we locate in the learned map also tracks time to failure across configurations, and one-step error does not. Failure is scored by a criterion that detects freezing as well as blow-up. An on-policy-only model holds its energy in band for 674 steps but stops moving at step 86, so energy alone would rank it beside the full method. \textbf{(C3) Gain and attribution.} On $128^{3}$ turbulence, time to failure rises from 21 steps to 721 on both criteria, at least $6.6\times$ that of the strongest method we compared, and the step-15 error is the lowest. The generators play complementary roles in addressing blow-up and freezing, and new clean DNS data at the labeled-frame count of the pool does not reproduce the gain. The benefit extends to a second regime and in part to a spectral backbone.

%% file: sections/2_problem.tex
\section{Diagnosing Long-Rollout Failure in Turbulence}\label{sec:problem}

\begin{figure}[t]
\centering
\includegraphics[width=0.97\linewidth]{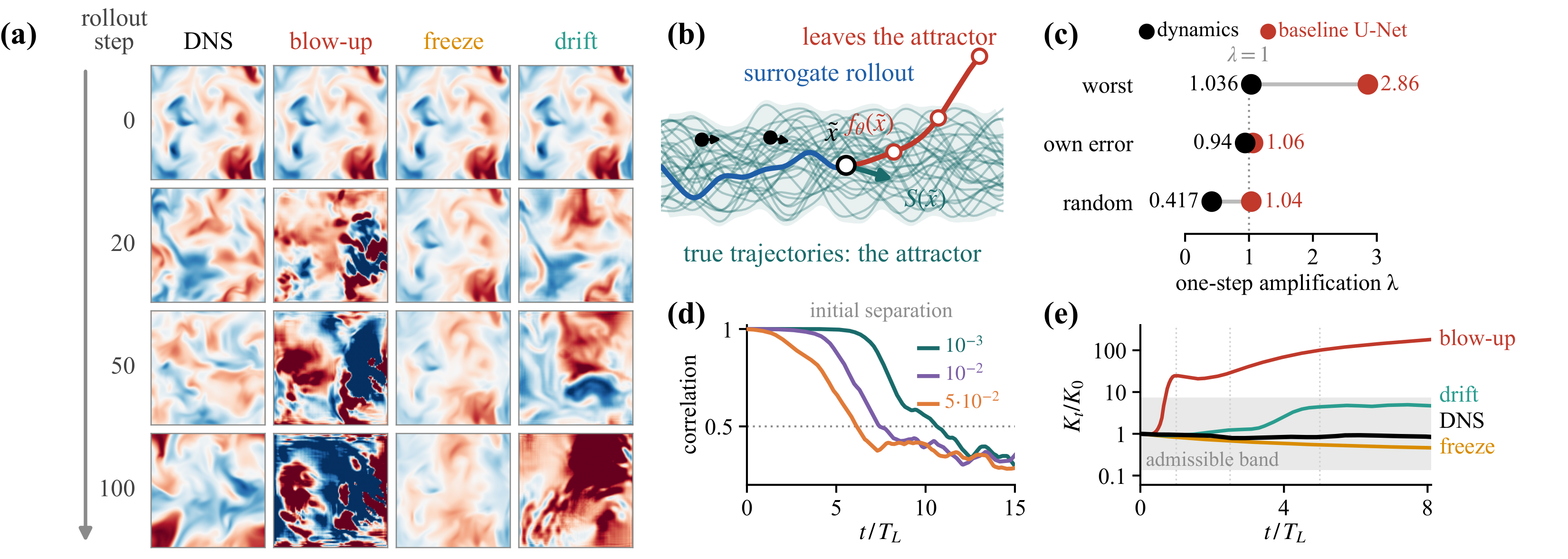}
\vspace{-0.25cm}
\caption{The three failure modes, and the one-step response behind them (Sections~\ref{sec:setting}--\ref{sec:markov}). (a) Mid-plane $u_x$ of a held-out DNS trajectory and of three rollouts of models trained with clean-future labels, one column each and one row per rollout step. (b) A rollout leaving the attractor. (c) One-step amplification in the dynamics and in the learned map. (d) Divergence of two nearby DNS trajectories. (e) Total energy of the columns of (a). Dotted: the steps of (a).}
\vspace{-0.14cm}
\label{fig:problem}
\end{figure}

\subsection{The attractor, and how rollouts leave it}\label{sec:setting}

This study examines forced homogeneous isotropic turbulence. Energy enters at the large scales, the nonlinear term carries it toward smaller ones, and viscosity removes it at the smallest resolved scales \citep{pope2000turbulent}. Once statistically stationary, the flow visits a particular set of velocity fields over which energy input and dissipation balance on average, which we call the \emph{attractor} of the flow. A simulation dataset draws only from that set, so it constrains a surrogate only on the states it contains (Fig.~\ref{fig:problem}b). At inference the surrogate is applied to its own output, on states no training trajectory visits. Throughout, $f_\theta$ is the surrogate's one-step map and $S$ is one step of the reference DNS, deterministic once the forcing realization is recorded (Appendix~\ref{app:twin}). One step advances the flow by $0.05\,T_L$, where $T_L$ is the large-eddy turnover time, and we report times in $T_L$. States taken from DNS trajectories are \emph{clean}, a \emph{rollout} applies $f_\theta$ repeatedly with no further input, and a predicted state is \emph{admissible} if it could be a state of the flow.

A rollout becomes inadmissible in one of three ways (Fig.~\ref{fig:problem}a,e). (1) In a \emph{blow-up}, the energy grows instead of settling, as the baseline U-Net of Fig.~\ref{fig:problem}e does within a turnover time. (2) In a \emph{freeze}, the learned map collapses toward the identity while the energy stays in its physical range, the shaded band of Fig.~\ref{fig:problem}e (Section~\ref{sec:protocol}). This is equally unphysical, since the energy cascade is continuous in time and a map that reproduces its input transfers no energy between scales. The clean-trained FNO freezes, and its frames stop changing while its energy decays inside the band. (3) In \emph{statistical drift}, the state keeps moving inside the energy band, but its spectrum and enstrophy depart from the reference. The third rollout in the figure, a U-Net fine-tuned on off-attractor states with clean-future labels (Section~\ref{sec:markov}), drifts to twenty times the reference enstrophy while its energy stays in band.

\subsection{The predictability horizon, and what chaos does not explain}\label{sec:horizons}

Turbulence is chaotic. Two DNS trajectories that start $0.1\%$ of the velocity norm apart under the same forcing separate with an e-folding time of $1.15$ turnover times (Appendix~\ref{app:twin}). That rate and a chosen tolerance set a \emph{predictability horizon} of several turnover times, beyond which pointwise agreement is lost \citep{boffetta2002predictability}. On a held-out initial condition the correlation of such a pair stays above $0.9$ for seven turnover times and falls below $0.5$ after ten (Fig.~\ref{fig:problem}d). Starting $5\times10^{-2}$ apart, the size of the baseline's one-step error, the pair needs six turnover times to reach $0.5$, long after the blow-up and freeze rollouts of Fig.~\ref{fig:problem}e have failed. A model that minimizes pointwise error beyond the horizon is rewarded for predicting the conditional mean of the futures its input still allows, which collapses their spread onto one field \citep{dai2026flowlearners}. Beyond it a rollout can only be asked to stay admissible with the right statistics. Chaos does not, however, explain the failures of Section~\ref{sec:setting}. The clean-trained baseline fails within about one turnover time, roughly one e-folding time. Over that time chaos multiplies a small discrepancy by $e$, so a $5\%$ discrepancy would grow to about $14\%$, whereas a rollout that blows up has by then multiplied its energy by $e^{2}$. Section~\ref{sec:markov} measures the response of the dynamics to a deviation of the kind a surrogate makes.

\subsection{Measuring how the dynamics and the learned map respond to a deviation}\label{sec:markov}

A rollout's error is a deviation from the attractor, and its evolution over the next step depends on which map advances it. Under a recorded forcing realization, one DNS step is the map $S$ of Section~\ref{sec:setting}. It is Markovian, since the next state depends only on the current state and the forcing, and it is defined at a deviated state as at any other (Appendix~\ref{app:twin}). Let $x$ denote a clean state and $\tilde{x}$ a deviated one. The Markov property fixes the training pair at $\tilde{x}$, whose future is $S(\tilde{x})$ (Fig.~\ref{fig:problem}b). Noise injection, the pushforward trick, and unrolled training (Section~\ref{sec:intro}) instead pair it with the clean future $S(x)$ of the state it came from, the \emph{clean-future label}. That label is off by $S(\tilde{x})-S(x)$, the response of the dynamics to the deviation. We call a method using this label a \emph{clean-label method}. This response is measurable. We write $\lambda$ for the factor by which one step multiplies a small perturbation of a given state along a given direction, and $\lambda^{\mathrm{model}}$ for the same quantity with $f_\theta$ in place of $S$. Both are measured at the same relative amplitude, $10^{-3}$ of the velocity norm, so that the two maps are compared on the same perturbation.

An isotropic Gaussian perturbation, whose energy lies at the smallest resolved scales where viscosity acts fastest, gives $\lambda=0.417$, so one step removes most of it. The worst direction we found, by power iteration through the DNS (Section~\ref{sec:generators}), gives $1.036$, a single-step gain and not an asymptotic rate (Table~\ref{tab:probes}). A surrogate's own error, whose energy lies mostly at the large scales where the signal is (Fig.~\ref{fig:longhorizon}d), gives $0.94$, and a perturbation shaped like the energy spectrum itself gives $1.016$. The dynamics therefore damps small-scale deviations and leaves large-scale ones nearly unchanged over one step, which sets how far the clean-future label is from $S(\tilde{x})$. For a small-scale deviation such as injected noise, the clean future is nearly the label the Markov property requires. For a deviation of the kind a rollout produces, the two futures differ by about the deviation itself, so the label moves with the input and cannot be kept. Along the direction a surrogate's error occupies, one step of the dynamics neither removes it nor amplifies it, so whether it grows depends on the learned map.

The same probes applied to $f_\theta$ show the learned map amplifying where the dynamics does not (Fig.~\ref{fig:problem}c). The baseline U-Net amplifies its worst direction by $\lambda^{\mathrm{model}}=2.86$ and carries small-scale content forward at $1.04$, where the dynamics removes it at $0.417$. Along its own error direction it amplifies at $1.06$, against $0.94$ for the dynamics. These directions are also not the physical ones. The cosine between a surrogate's direction of steepest growth and the worst direction of the dynamics never exceeds $0.004$ in absolute value (Appendix~\ref{app:extended}). Blow-up is therefore not inherited instability but amplification the map produces in directions no clean training pair constrains. Section~\ref{sec:mechanism} measures it for seven configurations, and Section~\ref{sec:method} trains at such states with $S(\tilde{x})$ as the label.

\subsection{Where the error spectrum exceeds the signal}\label{sec:spectra}

The energy of turbulence is concentrated at large scales and falls by orders of magnitude toward the smallest resolved scales, which carry the dissipation. A mean-squared error weights every mode by its absolute error, so modes orders of magnitude below the peak are effectively unsupervised, and networks also exhibit a low-frequency bias \citep{rahaman2019spectral}. This small-scale deficit has been addressed by spectral and scale-resolved losses and by iterative refinement \citep{chakraborty2026bsp,lippe2023pderefiner,kohl2023acdm,khodakarami2026spectral}. A clean-trained U-Net shows it, with a relative error that exceeds the signal itself near the dissipation scale and reaches $304\%$ (Fig.~\ref{fig:longhorizon}d, Appendix~\ref{app:extended}). The deficit shows in the statistics of a rollout rather than in when it fails. Off-attractor supervision does not address it, since the loss family is the baseline's throughout (Section~\ref{sec:label}). A late failure is therefore not sufficient evidence of fidelity, which is why the drift statistics of Section~\ref{sec:protocol} are reported with the failure times.

%% file: sections/3_method.tex
\section{Method: Off-Attractor Supervision (OAS)}\label{sec:method}

\begin{figure}[t]
\centering
\includegraphics[width=0.95\linewidth]{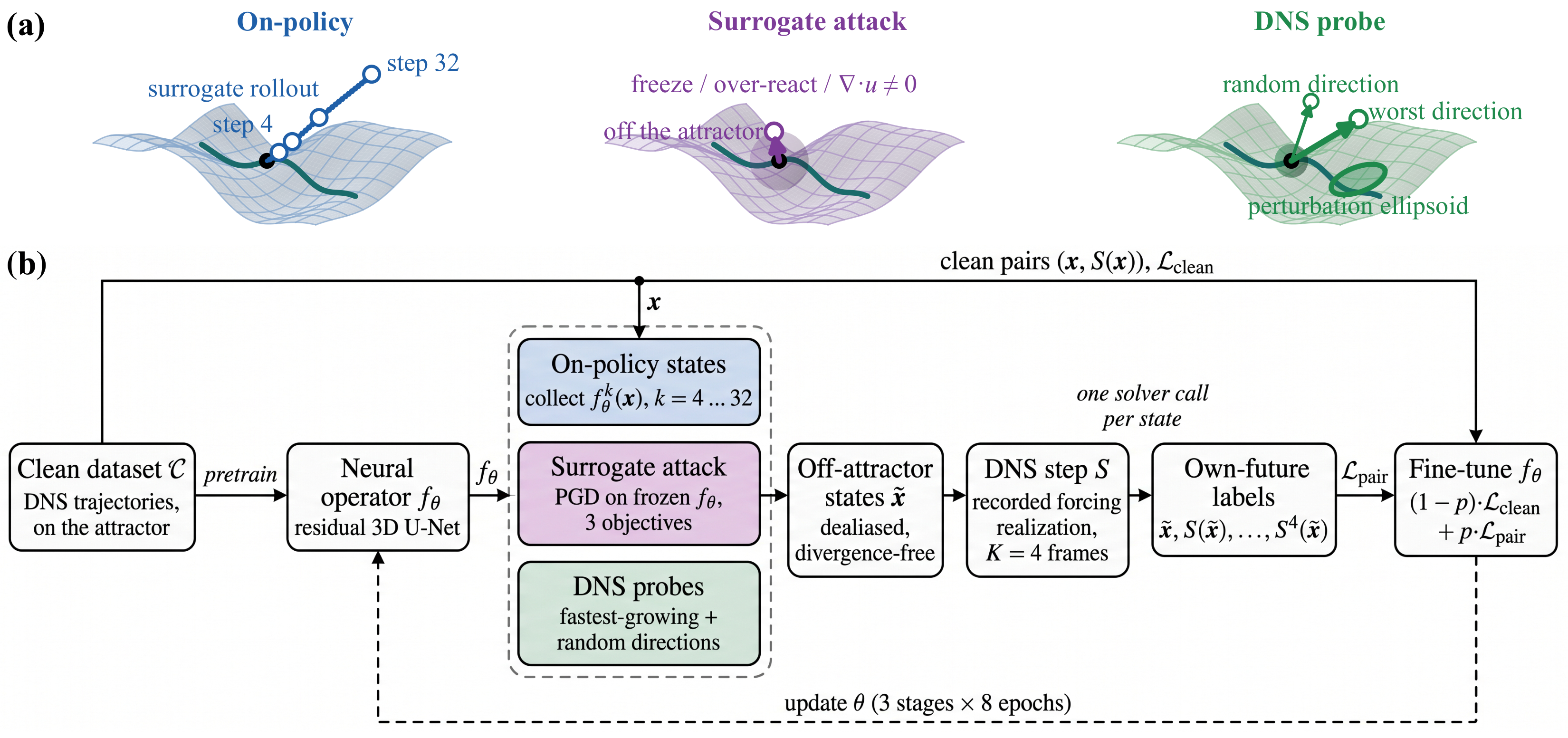}
\vspace{-0.1cm}
\caption{(a) The three state generators, drawn on the attractor. (b) The pipeline, from a generated state to a labeled pair in fine-tuning (Section~\ref{sec:training}).}
\vspace{-0.2cm}
\label{fig:method}
\end{figure}

\subsection{The surrogate and the reference map}\label{sec:setup}

We learn a deterministic one-step surrogate $u_{t+1}=f_\theta(u_t)$ of forced three-dimensional homogeneous isotropic turbulence and deploy it autoregressively, $u_{t+k}=f_\theta^{k}(u_t)$, for hundreds of steps from a single initial state without additional inputs. A state carries the three velocity components and a diagnostic pressure on a periodic grid, and the map is state-to-state. The surrogate therefore has neither the forcing realization nor a time index at inference. The reference dynamics is a resolved DNS, a double-precision pseudo-spectral solver with $2/3$-rule dealiasing and Eswaran--Pope stochastic forcing at the largest scales \citep{eswaran1988forcing}. It supplies the clean trajectories as well as the labels (Appendix~\ref{app:twin}). One step of it is the map $S$ of Section~\ref{sec:setting}, deterministic once the forcing realization and the time it is replayed from are fixed, and defined off the attractor as well as on it. We write $S^{j}$ for $j$ steps under the same forcing realization, so that $S(x),S^{2}(x),\dots$ are consecutive frames of one reference trajectory. Clean training data are the pairs a DNS trajectory supplies. The baseline minimizes the one-step error $\mathcal{L}_{\mathrm{clean}}(\theta)=\mathbb{E}_{x\sim\mathcal{C}}\|f_\theta(x)-S(x)\|^{2}$ over them, where $\mathcal{C}$ is the set of training states, all of which lie on the attractor (Section~\ref{sec:setting}). Section~\ref{sec:protocol} gives the dataset, backbone, and training budget.

\subsection{The own-future label}\label{sec:label}

Off-attractor supervision (OAS) gives a deviated state $\tilde{x}$ the \emph{own-future label}, its own next frame $S(\tilde{x})$ computed by the DNS under a recorded forcing realization (Fig.~\ref{fig:method}b). No clean future enters the label, and as $S$ is defined at a deviated state as at any other (Section~\ref{sec:setup}), the label is exact wherever the deviation lands, once the state is made solver-compatible (Section~\ref{sec:training}). Nothing else changes. The architecture and loss family are the baseline's, and the sample and gradient budgets are those of the comparison methods (Section~\ref{sec:protocol}). Two controls separate the label from the training schedule and the amount of new DNS data. Fine-tuning on clean windows gives 38 steps, and a matched budget of new clean DNS frames gives 47, versus 721 for own-future pairs (Section~\ref{sec:components}).

\subsection{Which states to relabel}\label{sec:generators}

Not every state a model might visit is worth relabeling, so three generators select states where the surrogate is likely to fail (Fig.~\ref{fig:method}a). Two of them generate their states without the solver, and the third uses it only to choose its two perturbation directions. The solver labels the states of all three.

\textbf{On-policy states.} We roll out the surrogate from training states and collect the states it reaches at depths of 4 to 32 steps, a range extending from a fraction of a turnover time to beyond one e-folding time (Section~\ref{sec:horizons}). These are the on-policy visits of imitation learning \citep{ross2011dagger}, with the DNS as the expert. They change as the model is retrained. Adding depths 64 and 96 to the full pool, however, brings the freeze back in every seed (Section~\ref{sec:components}), so the pool is limited to depth 32.

\textbf{Surrogate attacks.} The directions along which a model is fragile are not the ones that the dynamics amplifies (Section~\ref{sec:markov}), so the states on which it fails can be found by attacking the model alone, with no solver in the attack loop. Projected gradient ascent \citep{madry2018pgd} on the frozen surrogate perturbs a training state $x$ by $\delta$ to optimize one of three objectives, one per failure mode of Section~\ref{sec:setting}. \emph{Freeze-inducing} drives the map toward the identity, minimizing $\|f_\theta(x+\delta)-(x+\delta)\|$. \emph{Amplification-inducing} maximizes the local amplification $\|g_\theta(x+\delta)-g_\theta(x)\|/\|\delta\|$ of the predicted increment $g_\theta=f_\theta-\mathrm{id}$, the network's part of $\lambda^{\mathrm{model}}$ (Section~\ref{sec:markov}). \emph{Physics-violating} maximizes the two residuals a predicted state should not carry, the incompressibility residual of the output and the change it makes to the total energy (Appendix~\ref{app:settings}). The search runs thirty gradient steps per state at step size $0.1$. Between them the perturbation is projected back onto a divergence-free, dealiased sphere of fixed amplitude, $5\%$ of the norm of the state and the size of the baseline's one-step error (Section~\ref{sec:horizons}). The three objectives are complementary and are used together, in equal numbers. An objective that instead maximizes the one-step error, which puts the solver back inside the attack loop, is less effective (Section~\ref{sec:components}). The solver is called once per state, after the search, to compute the label. Conventional adversarial training on a PDE surrogate has no such separation, which is why it has been judged impractical even in two dimensions \citep{chen2026adversarialcfd}.

\textbf{DNS probes.} The two direction families that measure the dynamics in Section~\ref{sec:markov} also serve as generators. The first is the most amplified direction found by power iteration through the DNS. The second is the least amplified of twelve random isotropic divergence-free directions, whose energy lies at the small scales the dynamics removes. Applied to training states at the same amplitude as the attacks, they expose the model to both responses of the dynamics, the amplifying one and the contracting one. They are the pool's only model-independent states. The attacks cannot substitute for them, since a surrogate's steepest direction is nearly orthogonal to that of the dynamics.

\subsection{Relabeling and training}\label{sec:training}

We first make every generated state \emph{solver-compatible} by $2/3$-rule dealiasing followed by a projection onto the divergence-free subspace, since the DNS can only start from such a state. The term describes a requirement of the solver, and we reserve \emph{admissible} for the physical sense of Section~\ref{sec:setting}. The DNS then advances the projected state $\hat{x}$ for $K=4$ frames under a shared recorded forcing realization, so that one pair carries its future at four horizons, and the pressure is recomputed from the velocity. Attack and probe pairs take $\hat{x}$ as input. On-policy pairs keep the raw model output as input, so the model also learns to map its own unprojected output onto the solver's trajectory (Appendix~\ref{app:twin}). The horizon $K=4$ matches the four-step rollout of unrolled training (Section~\ref{sec:protocol}), so the two differ in their states and labels and not in how far they unroll.

Let $\mathcal{G}$ denote the generated states. A pair stores its input $y_{0}$, which is $\tilde{x}$ for an on-policy state and $\hat{x}$ otherwise, and the labels $y_{j}=S^{j}(\hat{x})$ for $j=1,\dots,K$. It is scored by a four-step unrolled loss,
\begingroup
\setlength{\abovedisplayskip}{7pt}
\setlength{\belowdisplayskip}{7pt}
\setlength{\abovedisplayshortskip}{0pt}
\setlength{\belowdisplayshortskip}{4pt}
\begin{equation}
\mathcal{L}_{\mathrm{pair}}(\theta;\tilde{x})=\frac{1}{K}\sum_{j=1}^{K}\frac{\big\|f_\theta^{\,j}(y_{0})-y_{j}\big\|^{2}}{\big\|y_{j}-y_{j-1}\big\|^{2}},
\label{eq:lpair}
\end{equation}
\endgroup
which is valid at every $j$ because $y_{j}$ is the future of the pair's own state at that horizon, and the gradient is taken through the whole chain of $j$ applications. The denominator is the change the dynamics makes at that step, which does not depend on $\theta$ and carries no gradient. It fixes the scale of each term, so that a model which returns its input scores one at the first horizon, whatever the amplitude of the deviation. The normalization also reweights the pool, since a pair the dynamics carries far in one step counts for less than one it barely moves. Pairs enter ordinary fine-tuning of the baseline model at a fixed fraction $p=0.3$ of each epoch's sample slots, the remaining slots keeping the one-step objective, so that training minimizes $(1-p)\,\mathcal{L}_{\mathrm{clean}}(\theta)+p\,\mathbb{E}_{\tilde{x}\sim\mathcal{G}}\,\mathcal{L}_{\mathrm{pair}}(\theta;\tilde{x})$. Each epoch has the length of a baseline epoch.

Because the on-policy generator follows the model, $\mathcal{G}$ is assembled in stages, and each stage trains for eight epochs. The baseline model is fine-tuned on a first pool of on-policy states with the one-step pair loss ($K=1$). The resulting model is fine-tuned on an on-policy pool re-collected against it together with the surrogate-attack pairs, and the DNS-probe pairs are added in a third stage. This gives the pool of 242 pairs behind the main model (132 on-policy, 66 surrogate-attack, 44 DNS-probe). A one-shot variant instead builds the whole pool against the baseline model, which lowers stability and seed consistency (Appendix~\ref{app:extended}). Appendix~\ref{app:settings} gives the schedule, and Table~\ref{tab:knobs} the sensitivity to other hyperparameters.

%% file: sections/4_experiments.tex
\section{Experiments}\label{sec:experiments}

\subsection{Protocol}\label{sec:protocol}

\textbf{Setup.}
The dataset is forced homogeneous isotropic turbulence on a $128^{3}$ grid ($\nu=0.0055$, $\mathrm{Re}_\lambda\approx 47$, $k_{\max}\eta=1.65$) from the validated DNS of Section~\ref{sec:setup} \citep{dai2026tide}, split by trajectory into training, validation, and test sets (Appendix~\ref{app:twin}). The surrogate is a residual three-dimensional U-Net \citep{ronneberger2015unet,cicek20163dunet} with 28M parameters, trained on the one-step squared error. We also run the method on the FNO, the backbone that keeps its energy in band the longest (Appendix~\ref{app:extended}). We roll every model out from three held-out initial conditions until it fails, so every failure time we report is observed rather than censored. Rollouts are not bit-reproducible, and we do not interpret differences below the resulting run-to-run floor (Appendix~\ref{app:repro}).

\textbf{Comparison methods.}
We re-implement every comparison method on identical data, backbone, and evaluation setup. Each starts from the same 24-epoch one-step baseline and trains for 24 more on its own objective, as ours does, at the same sample and gradient budget. The four are \emph{noise injection} \citep{sanchezgonzalez2020learning,stachenfeld2022learned}, the \emph{pushforward trick} \citep{brandstetter2022message}, \emph{unrolled training} with the gradient taken through a four-step rollout \citep{um2020solver,list2025differentiability}, and \emph{denoising refinement} (FlowRefiner, \citealp{dai2026flowrefiner,lipman2023flowmatching}). Denoising refinement leaves the one-step objective in place and cleans up each predicted state with a flow-matching sampler on the same backbone. Unrolled training appears in its purely unrolled form, since its solver-in-the-loop form needs a differentiable solver at this resolution. The control in its place puts the solver inside the state search instead of the training loop (Section~\ref{sec:components}). Appendix~\ref{app:settings} gives the settings for all four.

\textbf{Criteria and metrics.}
Let $K_t$ be the total kinetic energy at step $t$ and $K_0$ that of the initial state. The \emph{band-exit step} is the first $t\geq 1$ at which $K_t$ leaves $[e^{-2},e^{2}]\,K_0$, a band far outside the energy fluctuations of the DNS (Appendix~\ref{app:criteria}). Energy alone does not detect a freeze, so we also record the \emph{freeze step}, the first step at which the relative change $\|u_{t+1}-u_t\|/\|u_t\|$ stays below $3\%$ for three consecutive steps while the energy is in band. The \emph{strict failure step} is the minimum of the two. The freeze criterion is designed for deterministic maps and is lenient for a stochastic sampler. We report both failure steps throughout, and they disagree exactly when a model freezes before its energy leaves the band. Neither detects drift, which keeps the energy in band and the state moving. We measure drift by three statistics over the first 200 steps, the \emph{enstrophy ratio} and the \emph{log-spectral distance} against the DNS \citep{pope2000turbulent} and the \emph{spread ratio} against a DNS ensemble \citep{price2025gencast,mahesh2024hens}. Over the short horizon we report the normalized RMSE (nRMSE) against the DNS at the same step. Appendix~\ref{app:criteria} defines all four, fixes the windows they are read over, and reports sensitivity to both thresholds.

\input{tables/table1_main_comparison}
\subsection{Main result}\label{sec:main}

\begin{figure}[t]\centering\includegraphics[width=\linewidth]{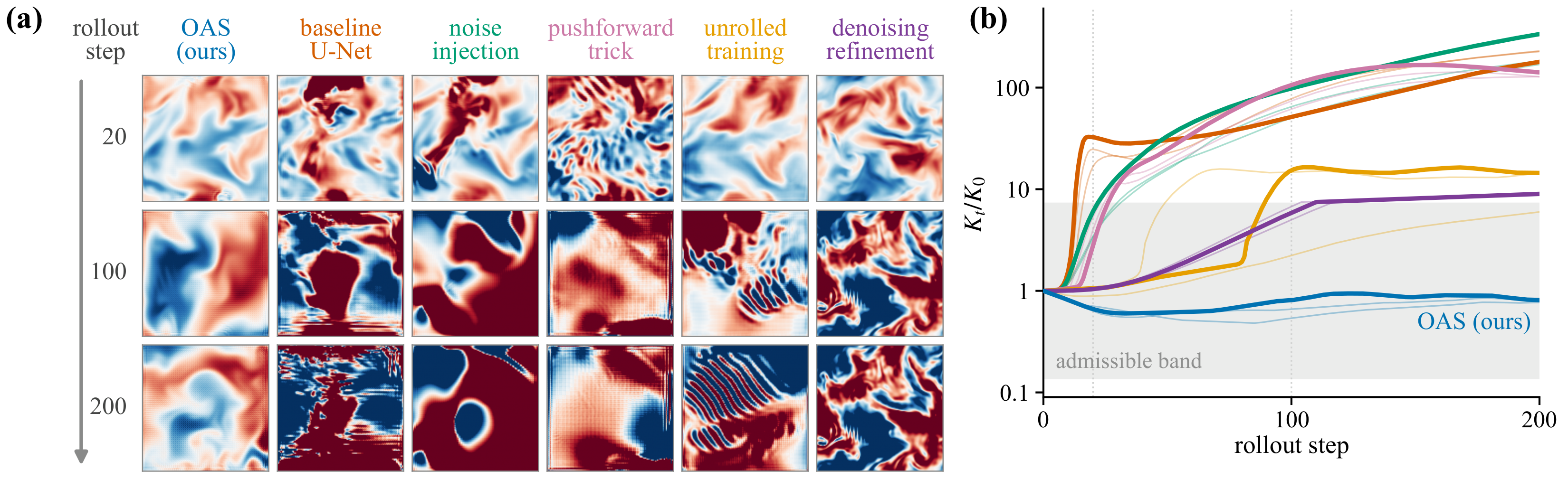}\vspace{-0.27cm}\caption{(a) Mid-plane $u_x$ of OAS and of the five other methods of Table~\ref{tab:main}, one column each and one row per rollout step, from one held-out initial state on the color scale of Fig.~\ref{fig:problem}a. (b) Total kinetic energy $K_t/K_0$, one color per method as in (a), the rollouts of (a) (thick) and two further held-out initial conditions (thin). Dotted: the steps of (a). Shaded: the band.}
\vspace{-0.02cm}
\label{fig:results}\end{figure}

\textbf{Time to failure rises from 21 to 721 steps.}
On both criteria, OAS raises the failure step of the baseline U-Net from 21 to 721. No other method in Table~\ref{tab:main} exceeds 110 on either criterion, so the gain over the strongest of them is $6.6\times$ on energy alone and $8.1\times$ on the strict criterion. Over five seeds OAS spans 705 to 784 with no freeze and no overlap with any comparison method (Table~\ref{tab:seeds}). Every comparison method ends in blow-up or freezing (Fig.~\ref{fig:results}). OAS also ranks best on every long-horizon metric of Table~\ref{tab:main}. The late-window enstrophy ratio falls from $932\times$ to $7.4\times$, and the ensemble spread reaches $0.91$ of the DNS ensemble against at most $0.45$ elsewhere.

\textbf{The benefit extends to a second regime and in part to a spectral backbone.}
A second regime at $\nu=0.010$ ($\mathrm{Re}_\lambda\approx 35$) has its own DNS dataset (Appendix~\ref{app:twin}) and 132-pair pool. One-shot OAS raises its strict failure step from 77 to 540 ($7.0\times$, band exit $7.6\times$, Table~\ref{tab:regime2}). With its own pool, the FNO improves from 2.120 to 1.141 in step-15 nRMSE and from 17 to 35 in strict failure step, but still freezes (Table~\ref{tab:oasbackbones}).

\subsection{Component ablations}\label{sec:components}
\vspace{-0.03cm}

\begin{figure}[t]
\centering
\includegraphics[width=\linewidth]{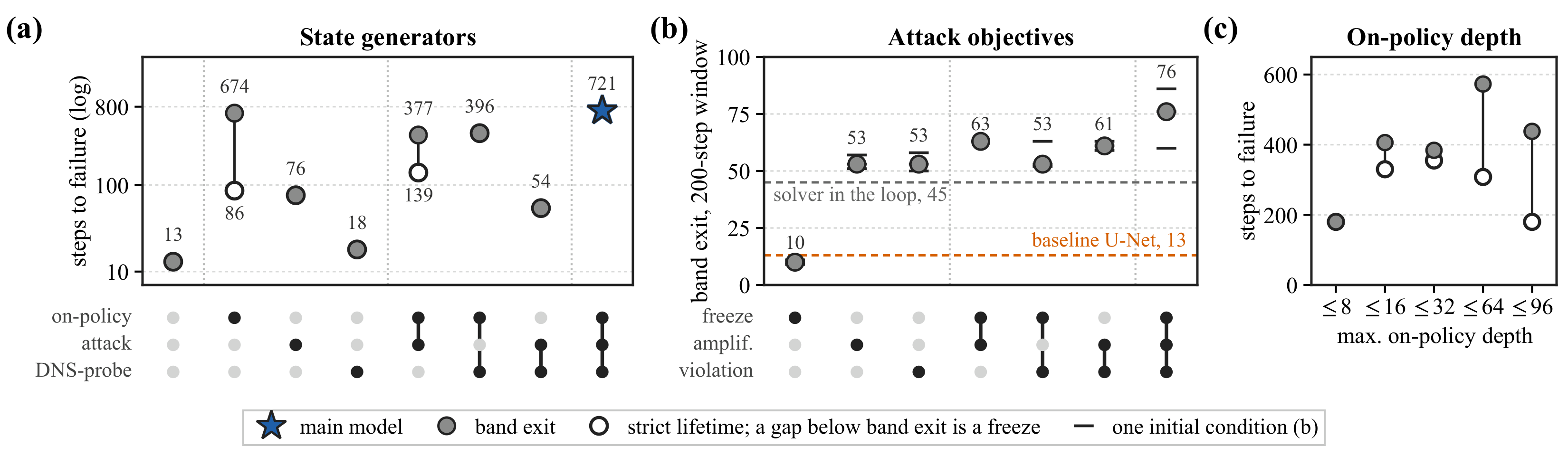}
\vspace{-0.73cm}
\caption{(a) The eight generator combinations, the matrix below marking which generators each pool holds. (b) The three attack objectives and their combinations, on attack pairs alone. (c) Maximum on-policy depth at matched budget.}
\vspace{-0.05cm}
\label{fig:ablations}
\end{figure}

\textbf{The three generators remove different failure modes.}
The on-policy-only stage connects OAS to imitation learning and earlier solver-based relabeling \citep{ross2011dagger,koehler2024apebench}. It removes blow-up but freezes at step 86 despite holding energy in band for 674 steps (Table~\ref{tab:ladder}). The two criteria diverge there, and on the strict criterion the stage and the full method differ by $8.4\times$. Adding surrogate-attack states delays freezing to step 139, with band exit at 377. DNS-probe states remove the freeze (Fig.~\ref{fig:ablations}a). Every generator is needed for the full gain: removing on-policy states reduces the strict failure step from 721 to 54, and removing surrogate-attack states reduces it to 396.

\textbf{The two failure modes trade against each other.}
No single attack objective suffices. Freeze-inducing pairs alone move the map away from the identity without limiting its increment, but cause earlier failure than the baseline. The other two objectives each reach 53 steps. Combining all three reaches 76 (Fig.~\ref{fig:ablations}b, Table~\ref{tab:attacks}), so the freeze-inducing and amplification-inducing objectives have to be combined. Maximizing one-step error, with the solver inside the attack loop, reaches 45. At matched budget, the strict failure step peaks at on-policy depth 32. Freezing rises from no initial conditions at depth 8 to all three at depths 64 and 96 (Fig.~\ref{fig:ablations}c, Table~\ref{tab:depth}). Freezing also returns with a lower pair fraction, fewer epochs, or a one-shot pool (Tables~\ref{tab:knobs},~\ref{tab:oneshot}).

\textbf{Own-future pairs drive the gain.}
With the same baseline and fine-tuning schedule, clean windows reach 38 steps, versus 524 on band exit and 173 on the strict criterion for one-shot OAS (Table~\ref{tab:labels}). At matched training and DNS budgets (984 clean frames versus 968 OAS labels), clean augmentation reaches 47 steps. This closes only about $1\%$ of the gap between clean-window fine-tuning and staged OAS. An unrolled-trained model fine-tuned on baseline attack pairs reaches only 99 steps, supporting model-specific state selection.

\begin{figure}[t]
\centering
\includegraphics[width=\linewidth]{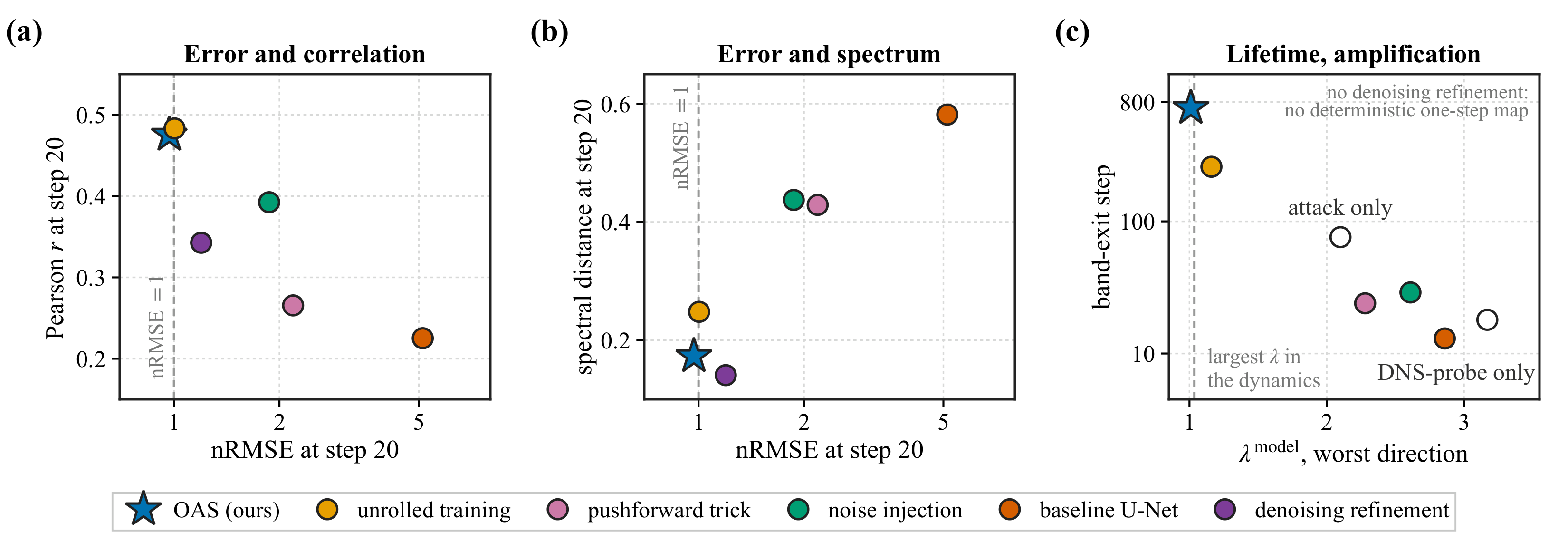}\vspace{-0.27cm}
\caption{Step-20 nRMSE versus (a) DNS correlation and (b) log-spectral distance for the six methods in Table~\ref{tab:main} (dashed: nRMSE of 1). (c) Band-exit step versus worst-direction amplification for seven deterministic configurations (dashed: the largest amplification found in the DNS).}
\label{fig:mechanism}
\end{figure}

\vspace{-0.05cm}
\subsection{What predicts failure}\label{sec:mechanism}
\vspace{-0.05cm}

\textbf{One-step error does not predict failure.}
All methods in Table~\ref{tab:main} use the same U-Net backbone, and deterministic rollouts initially track one another (Section~\ref{sec:protocol}, Figs.~\ref{fig:longhorizon} and~\ref{fig:mechanism}a). We report nRMSE at step 15 ($0.75\,T_L$), once rankings separate but before one e-folding time. OAS ranks best, despite the worst one-step validation loss among U-Net models ($1.7\times$ the baseline). Even the two longest-lived methods pass an nRMSE of 1 within about one turnover time (Appendix~\ref{app:extended}).

\textbf{Learned amplification tracks failure time.}
Across seven configurations, worst-direction amplification correlates with the band-exit step (Spearman $-0.93$, Fig.~\ref{fig:mechanism}c, Table~\ref{tab:lambdaconfigs}). Seven points are too few to establish a law. Every configuration above $2$ fails within 76 steps. OAS has the lowest amplification ($\lambda^{\mathrm{model}}=1.01$), is the only model near the DNS level ($1.036$), and fails latest on both criteria. Reaching that level takes states chosen against the model, and neither the attack nor the DNS-probe states get there alone. Random-direction amplification remains between $1.01$ and $1.04$ across configurations, versus $0.417$ for the DNS. OAS removes excess amplification but does not reproduce the contraction of the dynamics.

\textbf{Incompressibility violation does not explain the failure.}
As a control, we project every output of the baseline model onto the divergence-free subspace. On the baseline run of Table~\ref{tab:lambdaconfigs} this moves the band-exit step from 13 to 14, so the failure is not an accumulating violation of incompressibility.

\vspace{-0.1cm}

%% file: tables/table1_main_comparison.tex
\begin{table}[t]
\centering
\caption{Main comparison. Three held-out initial conditions, medians, over rollouts continued to failure. The baseline U-Net, unrolled training, and OAS are seed medians (Table~\ref{tab:seeds}). The other three methods are single runs. The three statistics are read over the first 200 steps ($10\,T_L$), the spread as a single reading at its end. Criteria in Section~\ref{sec:protocol}.}
\vspace{0.2cm}
\label{tab:main}
\footnotesize
\setlength{\tabcolsep}{4.5pt}
\begin{tabular}{lcccccc}
\toprule
 & nRMSE@15 $\downarrow$ & band exit $\uparrow$ & strict $\uparrow$ & enstrophy $\to\!1$ & spec.\ dist.\ $\downarrow$ & spread $\to\!1$ \\
\midrule
Baseline U-Net & 3.50 & 21 & 21 & 932$\times$ & 0.95 & 0.03 \\
Noise injection & 1.28 & 29 & 29 & 576$\times$ & 0.85 & 0.01 \\
Pushforward trick & 1.14 & 24 & 24 & 1771$\times$ & 0.97 & 0.02 \\
Unrolled training & 0.83 & 89 & 89 & 32.6$\times$ & 0.50 & 0.45 \\
Denoising refinement & 1.03 & 110 & 88 & 17.7$\times$ & 0.48 & 0.12 \\
\textbf{OAS (ours)} & \textbf{0.74} & \textbf{721} & \textbf{721} & \textbf{7.4$\times$} & \textbf{0.22} & \textbf{0.91} \\
\bottomrule
\end{tabular}
\end{table}

%% file: sections/5_limitations.tex
\vspace{-0.1cm}
\section{Limitations and Outlook}\label{sec:limitations}
\vspace{-0.1cm}

Both regimes are $128^{3}$ forced homogeneous isotropic turbulence at a low Reynolds number. The grid fixes it, since a resolved DNS needs $k_{\max}\eta\gtrsim 1.5$ and the dataset sits at $1.65$. Finer grids, other forcing, and flows with more structure in their attractor are the natural next settings. The gains are quantified in full on a convolutional backbone. On a spectral backbone, OAS roughly doubles the strict failure step and nearly halves the short-horizon error, but freezing persists (Appendix~\ref{app:extended}). Late-window statistics improve substantially but remain biased (Table~\ref{tab:main}). Failure time alone does not establish statistical fidelity. The learned map matches the dynamics along its worst direction, but along a random direction it still carries small-scale content forward where the dynamics removes it, and reproducing that contraction is an open problem. The own-future pairs also trade one-step accuracy for rollout accuracy \citep{yi2024tart}, and the one-shot variant varies across seeds. Isolating individual generators' causal effects requires matched training histories and budgets.

%% file: sections/6_conclusion.tex
\vspace{-0.1cm}
\section{Conclusion}
\vspace{-0.1cm}
In this paper, we proposed off-attractor supervision (OAS) for improving the long-term stability of neural surrogates for three-dimensional turbulence. OAS addresses insufficient supervision outside the training distribution through three state generators and offline DNS relabeling, pairing each selected state with its own physical future. Experiments show that OAS increases the median time to failure from 21 to 721 steps, reaching $8.1\times$ that of the strongest baseline under a criterion that detects both freezing and blow-up. Controlled experiments support the contribution of the selected states and their labels beyond changes in training schedule or additional DNS data. These findings motivate extending physically consistent supervision beyond reference trajectories as a strategy for stabilizing neural surrogates.

%% file: sections/appendix.tex
\section{Extended discussion of related work}\label{app:related}

Section~\ref{sec:intro} groups the stabilization literature by where a method acts, on the training inputs, on the predicted state, or on the learned operator. Three further distinctions organize the comparison drawn in this paper: which states the model is trained on, which future each of those states is labeled with, and where the reference solver is called, if it is called at all. Table~\ref{tab:related} arranges the families by those three.

\textbf{Deviated inputs with clean-future labels.}
Noise injection perturbs the input with a random field \citep{sanchezgonzalez2020learning,stachenfeld2022learned}, the pushforward trick perturbs it with the model's own one-step output and detaches the gradient through the perturbation \citep{brandstetter2022message}, and unrolled training exposes the model to the states its own rollout reaches over a short window \citep{um2020solver,kochkov2021ml,list2025differentiability,chakraborty2024multistep}. The three differ in how the deviation is produced but agree on what it is paired with. Noise injection adjusts its target so that the absolute one is unchanged: \citet{sanchezgonzalez2020learning} subtract the injected velocity noise from the acceleration target, so that integrating the prediction recovers the unperturbed next state, and \citet{stachenfeld2022learned} subtract it from the target increment to the same effect. The pushforward loss is $\mathcal{L}(f_\theta(u^{k}+\epsilon),u^{k+1})$, with $u^{k+1}$ read from the stored reference trajectory. No solver is re-run at the perturbed state in any of the three, and Appendix~\ref{app:settings} gives the implementations used here. The error that choice leaves is $S(\tilde{x})-S(x)$, and Section~\ref{sec:markov} measures it at both amplitudes that matter. For a small-scale perturbation, the dynamics removes most of the difference within one step, so the clean future is close to the correct label and the choice costs little. For a deviation of the size and spectrum a rollout actually produces, the two futures differ by about the deviation itself, so the label is wrong by roughly the quantity the model is being trained to handle.

\textbf{Selecting which states to train on.}
A second line chooses training points by a criterion rather than uniformly. \citet{ouyang2025rams} treat the sample locations as parameters and move them by ascent on the PDE residual, which needs no labels in the physics-informed setting and calls a solver at the moved inputs in the data-driven one. \citet{cesar2026ogas} train a generative sampler, though over configuration parameters rather than over states, and simulate fresh trajectories from the parameters it proposes, so the pairs it produces lie on the attractor. \citet{cao2026pinnadv} analyze a discriminator-based form of adversarial training for physics-informed networks, which is residual-based and needs no labels, so the difficulty discussed here does not arise there. The surrogate-attack and DNS-probe generators of Section~\ref{sec:generators} belong to this line and share its premise, that a labeling budget is better spent where a model is weak than spread over the attractor.

\textbf{Solver-computed labels.}
Three methods pair a deviated state with the solver's answer at that state, which is the labeling rule of Section~\ref{sec:label}. \citet{koehler2024apebench} define a \emph{diverted chain} objective in which the reference simulator branches off after each network prediction, so the target at a state the model rolled out to is the simulator advanced from that same state. The setup requires that simulator to be differentiable, and the benchmark reports it up to $32^{3}$. \citet{sun2025solver} states the distinction explicitly at adversarially found states, contrasting a loss against the solver's answer at the perturbed input with one against the clean answer and ablating the two, with the solver inside the attack loop and differentiated through, at about one solver forward and backward per attack step, on one- and two-dimensional problems. \citet{roy2026beyonduniform} select perturbed inputs by a gradient-free attack and label them with the solver at those inputs, on one-dimensional Burgers. Three further methods are adjacent without sharing the rule. Solver-in-the-loop places a differentiable solver inside the training loop so that its response to the corrected state enters the gradient, and keeps the reference trajectory as the target, so its deviated states carry clean-future labels \citep{um2020solver}. \citet{liu2025inverseevolution} build pairs through the evolution operator, consistent by construction and independent of any model. \citet{suriyanarayanan2026nudged} record the synchronization forcing of a nudged coarse run and learn it offline as a closure term rather than as a surrogate.

Off-attractor supervision differs from the first three in where the solver runs. It calls a non-differentiable DNS once per selected state, offline, after the states have been chosen, so that neither the attack nor the rollout collection touches it. The choice of states is then unconstrained by the cost of labeling, which admits a generator needing only gradients of the network (Section~\ref{sec:generators}), and the procedure reaches three-dimensional turbulence at $128^{3}$, where the three works above stop at $32^{3}$ or at two dimensions. Grid size is not itself the obstacle, since \citet{suriyanarayanan2026nudged} train offline against $128^{3}$ DNS for a closure term. The number of solver calls is. Fixing the pool between stages rather than refreshing it at every step is the cost of the arrangement. Section~\ref{sec:training} offsets part of it by re-collecting the on-policy states against the model the first stage produces, and Table~\ref{tab:oneshot} reports what a pool built in one pass gives up.

\textbf{Adversarial training of physics surrogates.}
\citet{chen2026adversarialcfd} audit FNO and U-Net surrogates on three two-dimensional flows and find them adversarially fragile. They also note why the usual remedy is hard to apply. A perturbed input changes the PDE solution, so a valid label needs a new simulation, which makes conventional adversarial training impractical for these surrogates. That is the mismatch of Section~\ref{sec:markov}, seen from the cost side, and it binds harder in three dimensions, where the resimulation is more expensive. The attack used here needs no solver. The three objectives of Section~\ref{sec:generators} are functions of the network alone, so the search costs forward and backward passes through it, and the solver is called once per state at the end, to supply the label.

\textbf{On-policy states and imitation learning.}
Collecting the states a policy visits and labeling them with an expert is \textsc{DAgger} \citep{ross2011dagger}, with the DNS as the expert. \citet{beatson2020composable} apply it in a PDE setting, querying a finite-element solver at the states a learned energy surrogate reaches. Their surrogate is static, so no question of which future to use arises. The first stage of Section~\ref{sec:training} is that algorithm, which is why Section~\ref{sec:components} reports it on its own: it holds the energy in band for 674 steps and stops moving at step 86, removing the blow-up mode and leaving the freeze. The limitation is structural rather than a matter of tuning. A rollout visits the states the model reaches, not the ones it would fail on and has not reached, and deeper rollouts do not fix this, since extending the pool to depths 64 and 96 makes the freeze more frequent rather than less (Table~\ref{tab:depth}). The generator ablation of Section~\ref{sec:components} attributes the removal of the freeze to the other two generators.

\textbf{Post-hoc correction and operator regularization.}
Two families act elsewhere and are in principle composable with this one. Post-hoc correction leaves the one-step objective in place and cleans up the predicted state, by iterative refinement or a diffusion or flow-matching sampler \citep{lippe2023pderefiner,kohl2023acdm,dai2026flowrefiner,yoo2026diffusionrollout}, or moves it back toward the attractor at inference \citep{pedersen2025thermalizer,liu2026selfrefining}. The last of these is the complement of what is done here: it returns a rollout to the set that training covers, rather than extending training to cover where rollouts go. Operator regularization acts on the learned map rather than on the data it is fitted on, by bounding the spectral norm of the convolution in the architecture itself \citep{mccabe2023stability}, by a spatially adaptive penalty \citep{nie2026jaws}, by penalizing the non-normality and non-commutativity of successive Jacobians \citep{pervez2026transient}, or by matching an invariant measure \citep{schiff2024dyslim}. The demonstrations are on two-dimensional flows, global weather fields, and chaotic systems of a few thousand state dimensions. This family addresses the failure mode of Section~\ref{sec:mechanism} by a different route. The penalties of \citet{pervez2026transient} constrain the \emph{structure} of the Jacobians, since normal and commuting Jacobians bound the gain of a product by the spectral radius rather than by the larger non-normal bound, whereas $\lambda^{\mathrm{model}}$ is a finite-amplitude magnitude. Neither quantity bounds the other. Here it falls from $2.86$ to $1.01$ with no term in the objective referring to it or to any Jacobian. Whether supplying the labels and constraining the map are redundant or complementary is not answered by any experiment reported here.

\textbf{Evaluating a long rollout.}
Reporting pointwise skill over a short horizon together with spectra and ensemble spread beyond it is established practice in operational weather and climate emulation \citep{lam2023graphcast,price2025gencast,bonev2025fourcastnet3,mahesh2024hens,wattmeyer2024ace2}, and Section~\ref{sec:protocol} follows it, with two additions specific to this setting. The failure time detects freezing as well as blow-up, since a map collapsing toward the identity keeps its energy in band indefinitely and is ranked well by an energy criterion alone, as the clean-trained FNO is in Table~\ref{tab:backbones}. Drift statistics are read beside it, since a late band exit leaves the statistics unconstrained, as the first training stage shows (Appendix~\ref{app:extended}). \citet{duraisamy2026predictivity} argues that part of what a surrogate of a multiscale system loses is not recoverable, since spectral bias and coarse-graining discard high-frequency content that no architecture or training procedure puts back. The deficit of Section~\ref{sec:spectra} is an instance of the spectral half of that argument. The failures studied here are of a different kind, occurring within a turnover time or two at the resolution of the reference itself, and Section~\ref{sec:markov} locates their cause in the learned map.

\begin{table}[!htb]
\centering
\caption{Where off-attractor supervision sits. \emph{States}: which inputs the model is trained on. \emph{Label}: which future each is paired with, $S(x)$ being the future of the clean state it was displaced from and $S(\tilde{x})$ its own. \emph{Solver}: where the reference solver runs, if at all. Rows are families rather than individual papers, and the text qualifies them. The own-future label appears in four of them, and the second and fourth columns are where those four differ.}
\vspace{0.2cm}
\label{tab:related}
\footnotesize
\newcolumntype{L}[1]{>{\raggedright\arraybackslash}p{#1}}
\setlength{\tabcolsep}{4pt}
\begin{tabular}{@{}L{3.2cm}L{2.9cm}L{2.8cm}L{3.5cm}@{}}
\toprule
\textbf{family} & \textbf{states} & \textbf{label} & \textbf{solver} \\
\midrule
One-step supervised & attractor only & clean future & offline, dataset only \\
\addlinespace[1pt]
Noise injection, pushforward & randomly perturbed & clean future $S(x)$ & none \\
\addlinespace[1pt]
Unrolled training & on-policy, short window & clean future $S(x)$ & none \\
\addlinespace[1pt]
Solver-in-the-loop & on-policy, solver-corrected & clean future $S(x)$ & differentiable, in the training loop \\
\addlinespace[1pt]
Diverted chain & on-policy & own future $S(\tilde{x})$ & differentiable, in the training loop \\
\addlinespace[1pt]
Solver-integrated attacks & adversarial & own future $S(\tilde{x})$ & in the attack loop, differentiated through \\
\addlinespace[1pt]
Active learning with solver labels & attack-selected & own future $S(\tilde{x})$ & online, once per selected input \\
\addlinespace[1pt]
Inverse-evolution augmentation & analytically constructed & consistent by construction & evolution operator, analytic \\
\addlinespace[1pt]
Post-hoc correction & attractor only & clean future & none, or at inference \\
\addlinespace[1pt]
Operator regularization & attractor only & clean future & none \\
\midrule
\textbf{OAS (ours)} & \textbf{on-policy $+$ attack $+$ DNS probe} & own future $S(\tilde{x})$ & \textbf{offline, once per selected state, not differentiated through} \\
\bottomrule
\end{tabular}
\end{table}

\section{Evaluation criteria, provenance, and sensitivity}\label{app:criteria}

\textbf{Definitions.}
Steps are counted from the first predicted frame (step 1). A rollout is continued until it fails, with no fixed evaluation horizon, so no failure time reported here is censored. In the freeze criterion of Section~\ref{sec:protocol}, the in-band condition prevents a blow-up from being recorded as a freeze through its inflated denominator. A stochastic sampler draws fresh noise at every step, so its relative change stays above the freeze floor even when its mean prediction has stopped moving, and the strict failure step of such a model is an upper bound. The failure-mode label in the tables records how a rollout ends. Within the first 200 steps the priorities are blow-up (energy exits upward or a non-finite value appears), then freeze. A rollout still in band and moving after 200 steps takes the label of its eventual terminal event, marked \emph{late} for the longest-running configurations here. Drift is not a terminal event. A rollout drifts when it stays in band and moving while the late-window energy or enstrophy ratio to the reference lies outside $[0.5,2]$ or the time-averaged log-spectral distance exceeds $0.5$, and the three statistics below report it. The enstrophy ratio divides the rollout's mean enstrophy over the late window, the last third of the 200 steps, by the stationary DNS value, the log-spectral distance is the relative $L^{2}$ distance between the logarithms of the shell-averaged energy spectra over the resolved shells, taken against the time-averaged reference spectrum, and the spread ratio divides the eight-member ensemble spread at a given step by the spread of the DNS ensemble started from the same eight perturbed states. None of the three depends on how far the DNS trajectory of an evaluation initial condition runs: they are a stationary value, a time average, and a separately integrated ensemble (Appendix~\ref{app:twin}). The pointwise error nRMSE at step $t$ is $\|u_t-u^{\mathrm{DNS}}_t\|_2/\|u^{\mathrm{DNS}}_t\|_2$, taken over the normalized state, the three velocity components and pressure. It reaches $1$ when the prediction is no better than the zero field and $\sqrt{2}$ for two decorrelated fields of matching variance. It is defined only while the reference exists, about 150 frames after an evaluation initial condition.

\textbf{Provenance of the band.}
Energy is quadratic in amplitude, so $K_t/K_0=e^{\pm 2}$ corresponds to a root-mean-square velocity one e-fold away from its initial value. On six long DNS records of about 125 turnover times each (Appendix~\ref{app:twin}), the energy fluctuates with a standard deviation of $12\%$ of its mean (per trajectory $10.5$ to $19.9\%$) inside an envelope of $[0.70,1.36]$ times the mean. The band edges lie about $7$ and $53$ standard deviations from the mean, and DNS trajectories never trigger the criterion.

\textbf{Sensitivity of the two thresholds.}
The freeze threshold beside it is fixed from the same records: the DNS changes by $10$ to $25\%$ of the velocity norm per frame, and the $3\%$ threshold lies a factor of about three below the slowest of those. Sweeping it needs no retraining, since the strict failure step can be recomputed from the stored per-step change rates. Over $\varepsilon\in\{1,2,3,5\}\%$ and persistence $k\in\{3,5,10\}$ steps, the main model's strict failure step equals its band-exit step at every cell with $\varepsilon\leq 3\%$, at any $k$: no seed freezes anywhere in that range, which is the criterion-independent form of the statement in Section~\ref{sec:components}. Its ratio to the first training stage stays between $4.3\times$ and $8.4\times$ across the grid and takes the value quoted in Section~\ref{sec:components} at the setting fixed in advance. Only at $\varepsilon=5\%$, which counts a state still moving by a twentieth of its norm each step as frozen, does the main model register a freeze, at step 516, and the ratio is then $8.3\times$. The variant carrying depths 64 and 96 is what the grid separates: its strict failure step falls by two fifths over the same range, from the band-exit step at $\varepsilon=1\%$ to $479$ at $\varepsilon=5\%$. Rescaling the band to $[1/b,b]\,K_0$ for $b\in\{1.5,2,3,5,10\}$ and recomputing every configuration's band-exit step leaves the ranking unchanged for $b\geq 2$. The value $b=1.5$ is excluded because it lies close to the observed envelope and produces false positives on rollouts that remain admissible. For a single seed, the ratio of the main model's band-exit step to that of unrolled training ranges from $2.9$ to $8.5$ over $b\in[2,10]$, and the value at $b=e^{2}\approx 7.4$ lies near the lower end of that range. The band preregistered before the experiments was $[0.2,5]\,K_0$, under which the same two methods give 770 and 239 steps ($3.2\times$). The switch to the $e^{2}$ band was made for its provenance and is conservative. Energy is also the last of the checks to trigger. The baseline model violates gradient-statistics and incompressibility checks within its first four steps, so the band-exit step is a late estimate of the failure time, not an artifact of a sensitive threshold.

\section{DNS, forcing, and dataset}\label{app:twin}

\textbf{Solver.}
The reference dynamics is an incompressible pseudo-spectral Navier--Stokes solver in double precision on a triply periodic $128^{3}$ grid with $2/3$-rule dealiasing, low-storage third-order Runge--Kutta time stepping with the viscous term integrated exactly by an integrating factor, and Eswaran--Pope stochastic (Ornstein--Uhlenbeck) forcing \citep{eswaran1988forcing} confined to the lowest wavenumber shell ($k<2$, correlation time $2.0$, variance $0.0555$). The corpus trajectories use a CFL-adaptive time step (CFL number $0.4$, $\Delta t\leq 0.008$) and export a frame at the first step past each multiple of $0.05\,T_L$ ($T_L=2.47$). Relabeling uses the same scheme at a fixed $\Delta t=3.98\times 10^{-3}$, so one model frame is 31 solver steps. Every label is driven by one recorded forcing realization per time-step size, started from a zero forcing state and replayed from its first step, so all pairs share the same forcing. The forcing state therefore reaches about $34\%$ of its stationary amplitude after one frame and $62\%$ after four, whereas the clean trajectories are sampled after a spin-up under stationary forcing. Replaying the recorded realization from the same state reproduces the trajectory with zero difference. Any state injected into the solver (on-policy, surrogate-attack, or DNS-probe) is first dealiased and projected onto the divergence-free subspace, and the pressure is recomputed from the velocity by the pressure Poisson equation when frames are exported. Surrogate-attack and DNS-probe pairs store that projected state as their input, so the model is trained at exactly the state its label was computed from. On-policy pairs store the raw rollout state, velocity and pressure as the model produced them, so they also supervise the projection. Collected states with large velocities are relabeled with a reduced time step ($\Delta t/m$, $m\leq 8$) to keep the CFL number below $0.45$, and states beyond that limit are discarded.

\textbf{Validation.}
The production dataset passes a validation protocol: $k_{\max}\eta=1.65$ (mean over its 2553 frames, with $99.9\%$ of the dissipation resolved), monotone spectral tail, box-to-integral-scale ratio $5.3$, stationarity drift $0.024\%$ per turnover time, energy-budget closure $0.07\%$, component and gradient isotropy within $1\%$ and $5\%$, incompressibility residual $\langle(\nabla\!\cdot\!u)^{2}\rangle/\langle|\nabla u|^{2}\rangle=10^{-29}$, velocity-derivative skewness $-0.51$, and a momentum-residual check of the Navier--Stokes equations on stored triples of frames with second-order convergence under time-step halving. Six long $256^{3}$ records of about $125\,T_L$ supply independent validation at higher $k_{\max}\eta$ and the energy-band statistics of Appendix~\ref{app:criteria}. They enter neither training nor evaluation. The second regime ($\nu=0.010$) passes the same checks ($\mathrm{Re}_\lambda=35.0\pm 1.3$, $k_{\max}\eta=2.59\pm 0.06$).

\textbf{Dataset.}
Sixteen DNS trajectories with independent forcing realizations make up the corpus. Each is spun up for 30 time units (about 12 turnover times) and then sampled at $0.05\,T_L$ for about eight turnover times, some 160 frames. The split is trajectory-atomic, eleven trajectories for training, two for validation, and three for evaluation, and normalization statistics are computed on the eleven training trajectories only. This gives about 1800 training frames, or 450 stride-4 rollout windows. The three evaluation initial conditions are frames of the three test trajectories, which span the same window as the training ones. The dispersion reference and the eight-member DNS ensemble behind the spread ratio at step 200 are produced separately, by re-integrating the eight perturbed states from a test-trajectory frame for 200 frames under a shared forcing realization.

\textbf{Hidden forcing.}
The surrogate never sees the OU forcing. To measure what the unobserved realization costs, we start from frame 10 of the reference trajectory, replay the recorded forcing for 20 frames so that the forcing state has reached its stationary amplitude, and from that single state advance one frame (31 solver steps) three times: under the recorded realization and under two fresh realizations of the same OU process (same forcing state, independent increments). The three next states differ by $0.80$ to $0.89\%$ of the velocity norm (the two fresh ones by $0.76\%$), whereas the state itself moves by $13.7\%$ in that step, and energies agree to $0.3\%$. Two trajectories started $10^{-3}$ of the velocity norm apart under the same forcing realization separate with an e-folding time of about 23 steps, the $1.15\,T_L$ quoted in Section~\ref{sec:horizons}. On a held-out initial condition, eight trajectories under one forcing realization, one unperturbed and seven perturbed by isotropic Gaussian fields (dealiased and projected), give $1.1$ to $1.2\,T_L$ by a log-linear fit over the first two to three turnover times, at amplitudes $10^{-3}$ and $10^{-2}$ alike. At $5\times10^{-2}$ the growth departs from exponential within the first turnover time, so no e-folding time is fitted there (Fig.~\ref{fig:problem}d).

\section{Training and generator settings}\label{app:settings}

\textbf{Backbone and training.}
The surrogate is a residual three-dimensional U-Net with $2.83\times 10^{7}$ parameters that predicts the increment to the next frame from the current frame (four channels: three velocity components and pressure). The baseline model trains for 24 epochs with batch size 1 and gradient accumulation over 8 samples, learning rate $10^{-3}$ with cosine decay, on windows sampled with stride 4 along each trajectory. OAS fine-tuning keeps that batch size and accumulation. It fills a fraction $0.3$ of the sample slots of each epoch with pairs, trained on the change-normalized pair loss of Section~\ref{sec:method}, and the rest with clean samples, trained on the one-step squared error. Each epoch has the length of a baseline epoch, and the clean windows and the pairs are each cycled through a fixed permutation across epochs. The main model's three stages run eight epochs each, 24 in total, the first at learning rate $3\times 10^{-4}$ with $K=1$ and the other two at $10^{-4}$ with $K=4$, each with cosine decay. Unrolled training applies the four-step loss to clean windows from the same pretrained baseline, on its own schedule rather than the OAS fine-tuning schedule above. Noise injection adds Gaussian noise of standard deviation $0.01$ in normalized units to the input and subtracts it from the target increment, and the pushforward trick unrolls two steps and detaches the first. These follow the original descriptions: the pushforward loss is $\mathcal{L}(f_\theta(u^{k}+\epsilon),u^{k+1})$ with $u^{k+1}$ the reference state, and unrolled training compares each unrolled prediction with the reference state at that step. Denoising refinement wraps the same U-Net in a decoupled flow-matching sampler with two refinement levels and two Euler substeps per level, at noise scales log-spaced in $[10^{-3},10^{-2}]$, so one rollout step costs five forward passes with fresh noise and a reported rollout is a single sample path. Half of each training batch trains the base predictor at refinement level $k=0$, which maps an all-zeros input to the clean next frame conditioned on the previous frame, and the other half trains the flow velocity $v_\theta(x_\tau,x_{\mathrm{prev}},k)$ with target $-\epsilon$. Optimizer, batch size, and gradient accumulation are those of the deterministic methods, and the residual connection is disabled because the base branch predicts absolute frames. At rollout step 1 on the evaluation initial conditions the five deterministic methods lie between $0.049$ and $0.056$ nRMSE. Denoising refinement predicts absolute frames from an all-zeros input and is not comparable there.

\textbf{Generators.}
On-policy states are collected from rollouts of an intermediate fine-tuning stage at depths 4, 8, 16 and 32 steps and relabeled for four frames (the deep variant of Table~\ref{tab:depth} adds 64 and 96). Surrogate-attack states use 30 steps of projected gradient ascent (step size $0.1$) on the frozen surrogate with the perturbation held on the sphere of relative amplitude $0.05$ after dealiasing and divergence-free projection, one objective per state, at states of training trajectories only. The one-step-error objective of Section~\ref{sec:components} recomputes the solver's answer at every ascent step and treats it as a constant, so its gradient also runs through the network alone. The two residuals entering the physics-violating objective are the relative incompressibility residual of the output, $r_\nabla=\langle(\nabla\!\cdot\!u)^{2}\rangle/\langle|\nabla u|^{2}\rangle$, and the relative change the step makes to the total energy $E$. They enter with equal weight and no further scaling, the objective being $\log r_\nabla\big(f_\theta(x+\delta)\big)+\big|\log\big(E(f_\theta(x+\delta))/E(x+\delta)\big)\big|$. Table~\ref{alg:oas} gives the order in which the pool is assembled. DNS-probe directions are obtained by power iteration through the DNS at relative amplitude $10^{-3}$ (fourteen iterations) for the growing direction and, for the contracting direction, by drawing twelve isotropic Gaussian directions (dealiased and projected), advancing each through one frame of the DNS, and keeping the least amplified. The resulting states use relative amplitude $0.05$. The main pool holds 132 on-policy, 66 surrogate-attack (22 per objective), and 44 DNS-probe (22 per family) pairs.

\begin{table}[!htb]
\centering
\caption{Off-attractor supervision: pool assembly and staged fine-tuning. Inputs are the clean training states $\mathcal{C}$, the solver $S$ with its recorded forcing realizations, the baseline model $f_{\theta_0}$ trained on $\mathcal{L}_{\mathrm{clean}}$, the pair fraction $p$, and the label horizon $K$. The three stages of Section~\ref{sec:training} differ in which generators have been added, and the first also in its horizon and learning rate (Appendix~\ref{app:settings}). The on-policy states are re-collected once, against the model that the first stage produces.}
\vspace{0.1cm}
\label{alg:oas}
\small
\begin{tabular}{@{}r@{\hspace{1.2em}}l@{}}
\toprule
\multicolumn{2}{@{}l@{}}{\textbf{Input:} $\mathcal{C}$, $S$, $f_{\theta_0}$, $p$, $K$; \quad \textbf{Output:} $\theta_3$} \\
\midrule
1 & $\mathcal{G}\leftarrow\varnothing$ \\
2 & \textbf{for} stage $m=1,2,3$ \textbf{do} \\
3 & \quad \textbf{if} $m\leq 2$: $\mathcal{G}\leftarrow\big(\mathcal{G}\setminus\text{on-policy}\big)\cup\{f_{\theta_{m-1}}^{k}(x)\,:\,x\in\mathcal{C},\ k\in\{4,8,16,32\}\}$ \\
4 & \quad \textbf{if} $m=2$: $\mathcal{G}\leftarrow\mathcal{G}\cup\{x+\delta^{\star}(x)\}$, $\delta^{\star}$ from PGD on frozen $f_{\theta_{m-1}}$, one objective per state \\
5 & \quad \textbf{if} $m=3$: $\mathcal{G}\leftarrow\mathcal{G}\cup\{x+\epsilon v\}$, $v$ from power iteration, or least amplified of 12 random directions \\
6 & \quad \textbf{for} each state $\tilde{x}$ newly added to $\mathcal{G}$ \textbf{do} \\
7 & \quad\quad $\hat{x}\leftarrow\Pi_{\mathrm{div}}\big(\mathrm{dealias}(\tilde{x})\big)$ \hfill \emph{solver-compatible start} \\
8 & \quad\quad store $\big(y_{0},\,S(\hat{x}),\dots,S^{K}(\hat{x})\big)$, $y_{0}=\tilde{x}$ if on-policy, else $\hat{x}$ \hfill \emph{one solver call per state, offline} \\
9 & \quad $\theta_m\leftarrow$ 8 epochs of $(1-p)\,\mathcal{L}_{\mathrm{clean}}(\theta)+p\,\mathbb{E}_{\tilde{x}\sim\mathcal{G}}\,\mathcal{L}_{\mathrm{pair}}(\theta;\tilde{x})$ from $\theta_{m-1}$ \hfill \emph{Eq.~\ref{eq:lpair}} \\
\bottomrule
\end{tabular}
\end{table}

\section{Extended results}\label{app:extended}
The figures and tables of this section back the statements of Section~\ref{sec:experiments} and follow its order.

\begin{figure}[!htb]\centering\includegraphics[width=\linewidth]{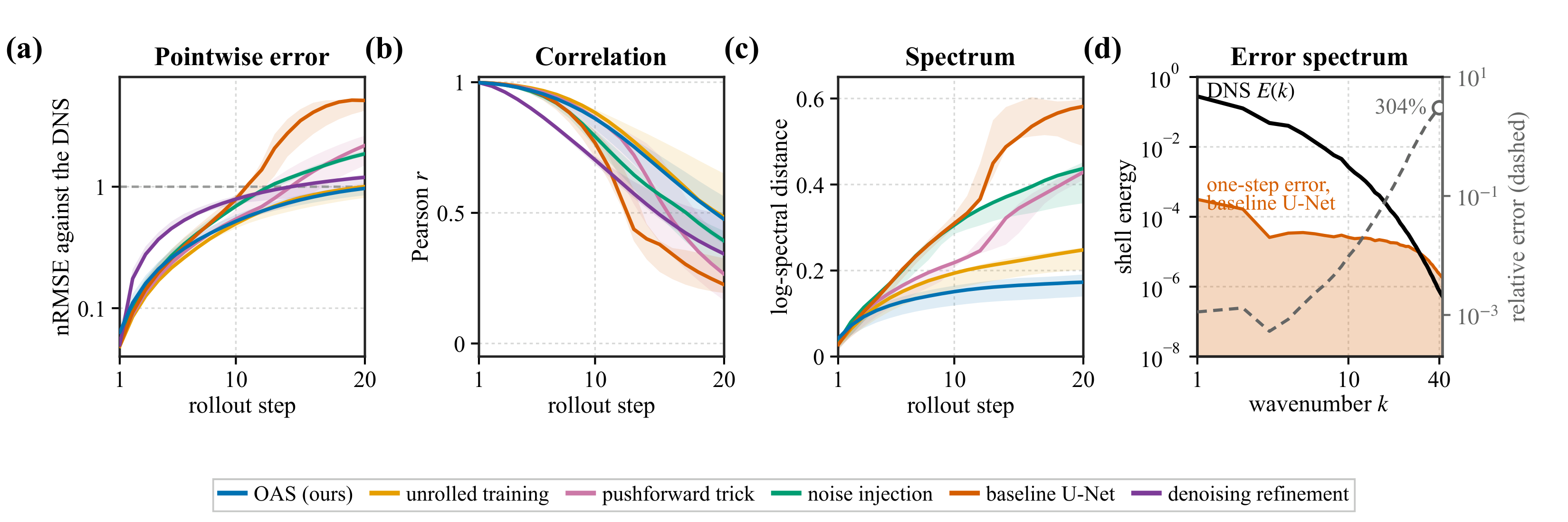}\caption{(a--c) Methods of Table~\ref{tab:main} over steps 1 to 20, medians over three initial conditions (shaded: min to max): nRMSE against the DNS (dashed: 1), Pearson correlation, and log-spectral distance, with denoising refinement omitted from (c). (d) DNS energy spectrum $E(k)$ (black, time average), shell energy of the baseline U-Net's one-step error (filled), and their ratio (dashed, right axis).}\label{fig:longhorizon}\end{figure}

\textbf{Short rollout and error spectrum (Fig.~\ref{fig:longhorizon}).}
Panels (a) and (b) follow the six methods of Table~\ref{tab:main} over the first twenty steps, (c) all but denoising refinement, and Fig.~\ref{fig:mechanism}a,b is the cross-section of (a) to (c) at step 20. At step 20 OAS has the lowest pointwise error, and its correlation with the DNS field is within $0.01$ of unrolled training, the second-best method on that error. Every method but OAS and unrolled training has passed an nRMSE of 1 by step 15, and those two reach it near step 20, before one e-folding time of the dynamics has passed. At step 15 the error of unrolled training exceeds OAS's by $12\%$. OAS then plateaus near $1.3$ for as long as the DNS reference runs, just below the $\sqrt{2}$ level of two decorrelated fields with matching variance, while unrolled training ($2.3$) and the baseline U-Net ($10$) pass through that level as their energy inflates. Panel (d) is the measurement behind Section~\ref{sec:spectra}. The DNS spectrum falls by orders of magnitude from the forced shell to the dissipation scale, and the shell energy of the baseline U-Net's one-step error falls more slowly, so the relative error rises with wavenumber and reaches $304\%$ at $k=40$. Every step of a clean-trained rollout therefore injects small-scale content that the dynamics would dissipate.

\textbf{Energy trajectories and dispersion (Fig.~\ref{fig:results}).}
The energy trajectories show blow-up for the baseline U-Net and the input-perturbation methods within the first 200 steps, delayed blow-up for unrolled training, a freeze at step 88 for denoising refinement, and for OAS, beyond the window of Fig.~\ref{fig:results}b, a slow drift out of the band at step 721 with no freeze before it. Fig.~\ref{fig:results}a shows the corresponding slices of every method on one initial condition. The input-perturbation methods and the baseline U-Net saturate the color scale within two turnover times, unrolled training leaves the band at step 90 on that condition, denoising refinement barely changes between the last two rows, and OAS stays in it to the end of the 200-step window, carrying large-scale structure of the kind the DNS column of Fig.~\ref{fig:problem}a shows. Its slices nonetheless smooth as the rollout proceeds, while its late-window enstrophy runs well above the reference (Table~\ref{tab:main}): the rollout has not failed, and its small-scale statistics have drifted, which is the distinction the drift metrics record. The DNS-ensemble dispersion agrees with the energy view. OAS overshoots the DNS ensemble spread by about $3\times$ at step 100 and returns to $0.91$ at step 200, the only method that converges back toward the reference. Unrolled training overshoots by $4\times$ and blows up, and the input-perturbation methods collapse to a few percent of the reference. A longer reading gives the growth of the spread rather than its level at one step. Eight members perturbed at $10^{-3}$ of the velocity norm along a spectrum-shaped direction and followed to step 700 on one initial condition disperse exponentially for all three models measured, with a log-linear e-folding time of $1.1\,T_L$ for unrolled training, $1.6\,T_L$ for OAS, and $10\,T_L$ for the baseline model, against the $1.1$ to $1.2\,T_L$ of DNS pairs (Appendix~\ref{app:twin}). The rate does not separate the two survivors. Where the spread settles does. Over the last third of that window it holds at $0.62$ of the velocity norm for OAS and runs to $0.99$ for unrolled training as its energy inflates, while the baseline model's ensemble never leaves $0.09$, the dispersion of a map collapsing toward a single field. This is one initial condition and eight members, so we read it for the shape of the three curves.

\begin{figure}[!htb]\centering\includegraphics[width=0.8\linewidth]{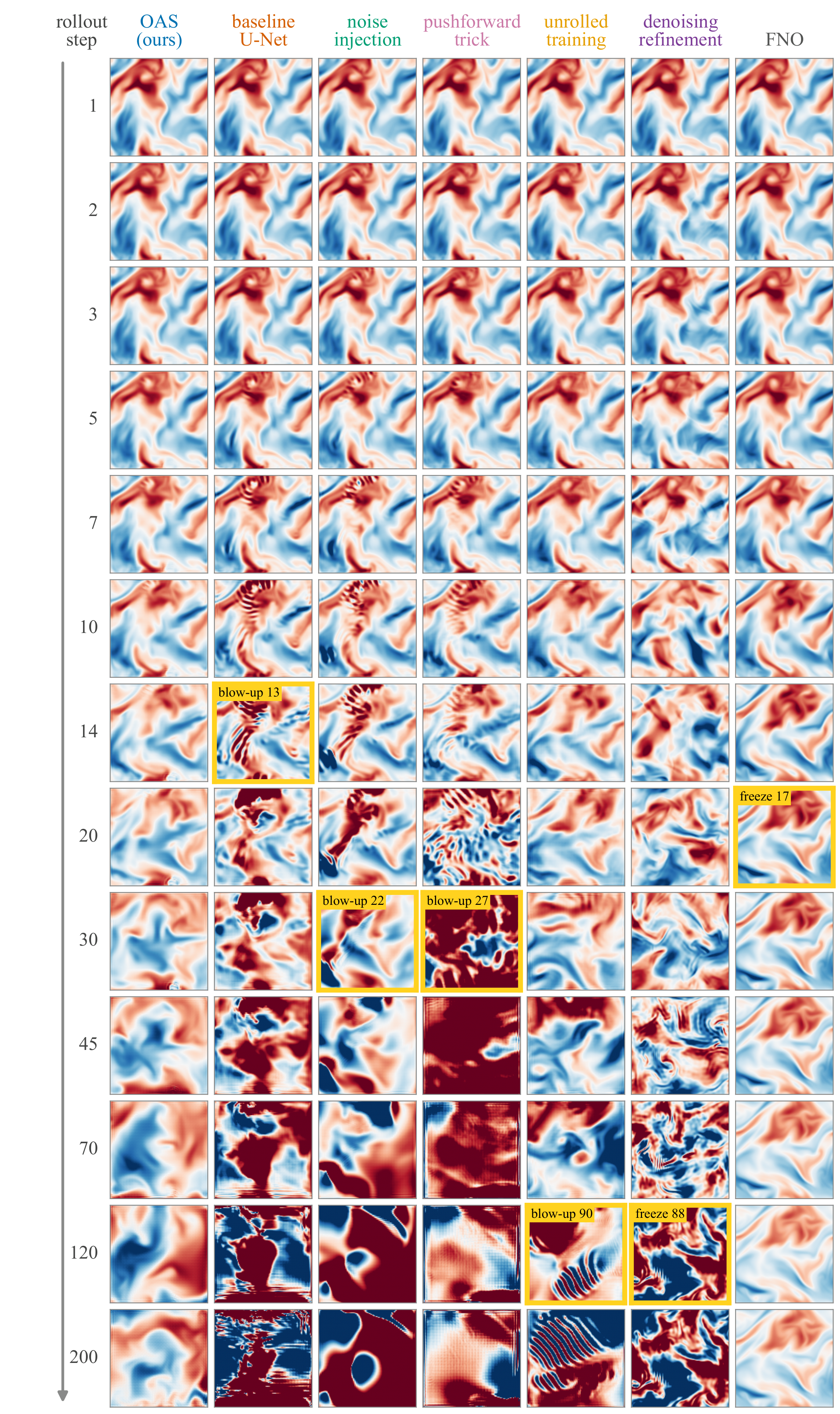}\caption{Onset of failure. Mid-plane $u_x$ of the six methods of Table~\ref{tab:main} and of the clean-trained FNO of Table~\ref{tab:backbones} at thirteen rollout steps, six of them in the first ten, one column each and one row per step, from one held-out initial condition and on the color scale of Fig.~\ref{fig:problem}a. Framed: the first step shown after the rollout fails, with its failure mode and failure step.}\label{fig:earlyfailure}
\vspace{-0.2cm}
\end{figure}

\textbf{The onset of failure falls between the rows of Fig.~\ref{fig:results} (Fig.~\ref{fig:earlyfailure}).}
Fig.~\ref{fig:results} samples three rollout steps, which is coarse where the deterministic methods break. The baseline U-Net is still at $K_t/K_0=1.01$ at step 5 and reaches $32.6$ by step 20, so its whole blow-up falls between two of those rows. Fig.~\ref{fig:earlyfailure} keeps its layout, adds the clean-trained FNO, and samples thirteen steps on one held-out initial condition, six of them in the first ten. The three methods that blow up early develop grid-scale structure within the first turnover time, while their large scales are still in place, which is the order the diagnosis of Section~\ref{sec:markov} predicts. The FNO freezes at step 17, and its field changes little after that. Off-attractor supervision and unrolled training hold their large-scale structure to step 45, and past it the two part, unrolled training into a striped pattern and off-attractor supervision into progressively smoother fields. This is one rollout of each method.

\textbf{Pool size (Table~\ref{tab:poolscaling}).}
Four pool sizes between 60 and 296 pairs, fine-tuned from the one-step baseline for eight epochs each, place the largest pool first on both criteria and above the smallest by $70\%$ on band exit and $55\%$ on the strict criterion. The two intermediate sizes do not fall between them, and the per-initial-condition spread is wide enough that the comparison carries only its endpoints. The reduced pools were drawn as random subsets of the pool carrying depths 64 and 96 rather than in the proportions of Section~\ref{sec:training}, so composition varies with size and the sizes are not comparable to the one-shot family below. In a single stage of eight epochs the largest of them reaches $79\%$ of the staged main model's strict failure step, which is the sense in which a few hundred states are enough. Across the same four, the one-step validation loss rises as the pool grows, so the endpoints repeat along this axis the inversion Section~\ref{sec:mechanism} reports across methods.

\input{tables/table15_pool_scaling}

\textbf{One-shot family (Table~\ref{tab:oneshot}).}
This family holds the base fixed and changes only the pool, fine-tuning the baseline model directly for the same 24 epochs on each reduced pool. Built in one pass this way the method is weaker than the staged main model and more seed dependent. Removing the on-policy states again collapses it, while removing the attack pairs or the two deepest depths leaves both criteria inside or at the edge of the five-seed range of the full pool (181 to 774 on band exit, 173 to 617 on the strict criterion), so in this setting, where configurations fail by freezing and its onset is seed dependent, those two contributions are not resolved. The staged training resolves all three.

\input{tables/table2b_oneshot_family}

\textbf{Generator combinations (Table~\ref{tab:ladder}, Fig.~\ref{fig:ablations}a).}
Table~\ref{tab:ladder} lists the eight combinations of Section~\ref{sec:components} with their freeze steps and failure modes. Rows 2 to 4 are the stages of the staged training, and row 5 adds depths 64 and 96 to the main pool. The strict criterion makes the sequence legible, since on energy alone the first stage (674) would place second in Table~\ref{tab:main}. The leave-one-out cells are built differently. Without DNS-probe states the model is the stage before they were added (139), whereas without surrogate-attack states or without on-policy states the final stage is retrained on the reduced pool (396 and 54). Measured on single runs, the first stage's late-window enstrophy is $48\times$ the DNS value and its log-spectral distance $0.42$, against $9.2\times$ and $0.24$ for one run of the main model, whose seed medians Table~\ref{tab:main} reports.

\input{tables/table2_component_ladder}

\textbf{Inside the attack (Table~\ref{tab:attacks}, Fig.~\ref{fig:ablations}b).}
The full $2^{3}$ grid of Section~\ref{sec:components} appears in Table~\ref{tab:attacks}, with per-initial-condition values and the solver-in-the-loop control. Each cell is fine-tuned from the same base on 66 pairs and read as the band-exit step, so the grid compares objectives at a fixed small budget. No configuration in the grid freezes, and the strict failure step coincides with band exit. The freeze-plus-amplification cell was evaluated on a single initial condition. The freeze-inducing objective is the instructive case. It costs three steps against the baseline's 13 when used alone, adds nothing beside the physics-violating objective (53), and adds fifteen on top of the other two (61 to 76). Pairs that only push the model away from the identity train it to move without constraining how far it moves, and the amplification-inducing pairs supply the bound. The solver-in-the-loop control maximizes the one-step error with about thirty solver calls per state. These are single-seed configurations and the pairs of objectives lie within a few steps of one another, so the grid is read for its pattern and not its ordering.

\textbf{On-policy depth and training hyperparameters (Table~\ref{tab:depth}, Table~\ref{tab:knobs}, Fig.~\ref{fig:ablations}c).}
At the matched budget of Table~\ref{tab:depth} (same base, 8 epochs, 176 pairs, with only the maximum rollout depth of the on-policy states varying) the band-exit view is inflated by freeze survivorship. The strict failure step peaks at a maximum depth of 32 (355) and declines as deeper states re-introduce freezing, with the freeze count rising from none at 8 to all three initial conditions at 64 and 96, while the two shallowest depths alone take the base from 377 to 180 on band exit and from 139 to 180 on the strict criterion. Table~\ref{tab:knobs} varies one hyperparameter at a time from the defaults of the staged main model (Appendix~\ref{app:settings}). Band exit falls by 15 to $40\%$ under the three changes and the strict failure step by 47 to $88\%$, since each change lets late freezing return.

\input{tables/table6_attack_grid}
\input{tables/table3_depth_ablation}
\input{tables/table7_recipe_knobs}
\FloatBarrier

\textbf{Label control (Table~\ref{tab:labels}).}
The two rows of Table~\ref{tab:labels} marked $\dagger$ share the baseline model and an identical fine-tuning schedule, the 24 epochs, the same learning rate, and the four-step unrolled loss of Appendix~\ref{app:settings}, so at a matched total budget of 48 epochs the training pairs are the only difference between 38 and 524 steps. The clean-window configuration also leaves the one-step validation loss at the base value. The 38-step and 259-step rows share their states, labels, loss, and base and differ only in schedule. The unrolled-training row is the single seed whose amplification Table~\ref{tab:lambdaconfigs} measures (259), and its three-seed median is 89 (Table~\ref{tab:seeds}). The matched-budget row controls for the amount of new DNS data. Six fresh forcing realizations supply 246 clean windows and 984 new frames, against the 968 the pool holds, at the $0.05\,T_L$ frame spacing the pair pool uses, and they enter the same pair recipe from the same base model on the staged schedule of the main model, so the control matches OAS in samples and optimizer steps as well as in DNS frames. It fails by blow-up on every initial condition, at 47 steps on both criteria (47, 50, 39 by initial condition). Clean data at the same budget therefore reaches nine steps beyond the clean-window row, about one percent of the 683 steps between that row and the full method.

\input{tables/table4_label_paradigm_cross}
\FloatBarrier

\textbf{Amplification (Table~\ref{tab:probes}, Table~\ref{tab:lambdaconfigs}).}
Table~\ref{tab:probes} lists the one-step gains of the dynamics along the four directions of Section~\ref{sec:markov}, all finite-amplitude single-step gains and not asymptotic rates. Table~\ref{tab:lambdaconfigs} lists the gains of the learned map for the seven configurations of Fig.~\ref{fig:mechanism}c, each measured on the single run whose failure times the table reports, so they differ from the seed medians of Table~\ref{tab:seeds}. Along a random isotropic direction every learned map stays between $1.01$ and $1.04$, and the ordering of the configurations is set by the worst direction alone. That direction is not the one the dynamics amplifies: the cosine of Section~\ref{sec:markov} stays below $0.004$, which is why the model-independent DNS-probe states alone leave the amplification above the baseline value ($3.17$) while the attack states alone bring it to $2.10$.

\input{tables/table8_lambda_probes}
\input{tables/table12_lambda_configs}
\FloatBarrier

\textbf{Backbones (Table~\ref{tab:backbones}).}
FNO and TFNO are spectral backbones, DPOT and FactFormer attention backbones, all trained on the one-step objective on the same data, every one a single run. All four carry at least twice the U-Net's one-step error, from $2.1\times$ for FNO to $3.3\times$ for DPOT, and fail within 6 to 19 steps on the strict criterion, so one-step error does not order the five by when they fail. Neither of the two error columns orders the other either. FactFormer is the more accurate of the two attention models at step 1 and the earlier of the two to fail, and both spectral backbones reach step 15 with less pointwise error than the U-Net whose one-step error is less than half of theirs, though every backbone is past the $\sqrt{2}$ level of two decorrelated fields by then. The two criteria separate FNO from the rest. It keeps its energy in band thirty times longer than the U-Net, 408 steps against 13, and has stopped moving by step 17. That combination is what spectral truncation would produce. Its spectral convolutions keep a fixed set of low wavenumbers, which can bias the map toward a smoothed field, and the large scales it retains hold the energy in band long after the map has stopped transferring any. A long band-exit step is therefore not stability here.

\input{tables/table5_backbones}

\textbf{Choosing the backbone of the main study (Table~\ref{tab:oasbackbones}).}
Table~\ref{tab:backbones} separates two criteria a backbone could be chosen by, and they disagree. The U-Net has the lowest error at one step, and the FNO keeps its energy in band the longest. Neither criterion is safe on its own. The U-Net's advantage at step 1 has inverted by step 15, where both spectral backbones are more accurate than it despite carrying two to three times its one-step error, and the FNO's 408 steps in band are those of a frozen map. We therefore ran the method on both candidates rather than settling the question by either number.

\textbf{Off-attractor supervision on a spectral backbone.}
The staged recipe of Section~\ref{sec:training} was applied to the FNO as it was to the U-Net, in the same three stages of eight epochs, with its own pool collected against it. Step-15 error and strict failure step each improve by about a factor of two, against $4.7\times$ and $34\times$ for the same method on the U-Net (Table~\ref{tab:oasbackbones}). The strict criterion is the one that reads this backbone, whose clean-trained band exit of 408 sits far beyond its freeze at 17, so energy alone measures a map that has already stopped moving. The failure mode is also unchanged. The FNO still freezes, later, whereas the U-Net under the same treatment freezes in no seed at all. The U-Net is the backbone of the main study for that reason, and not for its one-step error, which Section~\ref{sec:mechanism} shows does not order the rollout. It is the backbone the method has the most room to repair. The full gains are therefore quantified on the convolutional family, and a fraction of them carries to a spectral one.

\input{tables/table13_oas_backbones}
\FloatBarrier

\textbf{Second regime (Table~\ref{tab:regime2}).}
The dataset at $\nu=0.010$ passes the validation checks of Appendix~\ref{app:twin}. Its pool of 132 pairs (54 on-policy, 54 surrogate-attack, 24 DNS-probe) was built in one shot against that regime's baseline U-Net, and the fine-tune runs the same 24 epochs as in the main regime. The per-initial-condition values spread by 25 steps for the baseline model and by 86 steps for OAS on band exit.

\input{tables/table11_second_regime}
\FloatBarrier

\textbf{Seeds (Table~\ref{tab:seeds}).}
Over five seeds the main model spans 705 to 784 on both criteria, the variant with depths 64 and 96 added 685 to 833 on band exit and 583 to 718 on the strict criterion, and the one-shot variant 181 to 774 and 173 to 617. Staging therefore narrows the seed spread as well as raising the median. Unrolled training spans 52 to 259 over three seeds and the baseline model 13 to 22. Row 1 of Table~\ref{tab:ladder} is the single-seed baseline run at 13, whose three-seed median is 21 (Table~\ref{tab:main}).

\input{tables/table10_headline_robustness}
\FloatBarrier

\section{Reproducibility details}\label{app:repro}
Training and evaluation ran on two NVIDIA RTX 5090 (32~GB) workstations and, for five of the one-shot ablation configurations, on cloud A100 (40~GB) instances. Building and labeling the whole pool takes about as much wall-clock time as two of the 48 training epochs behind the main model, most of it the attack search on the network rather than the solver calls. The configurations of Table~\ref{tab:seeds} and the FNO under off-attractor supervision (Table~\ref{tab:oasbackbones}) were trained with several seeds, and the ablation configurations, the amplification probes, and the clean-trained backbones are single runs. Rollouts on GPUs are not bit-reproducible, which sets the evaluation noise floor: band-exit steps carry a run-to-run floor of 3 to 6\% and strict failure steps one of 6 to 11\%. Relabeling is bit-reproducible because it is driven by a recorded forcing realization at a fixed time step.

%% file: tables/table15_pool_scaling.tex
\begin{table}[!htb]
\centering
\vspace{0.15cm}
\caption{Pool size at a matched fine-tuning budget: the one-step baseline, eight epochs, and random subsets of the 296-pair pool that carries depths 64 and 96. Medians over three initial conditions with the per-condition strict values in brackets. Single runs.}
\vspace{0.15cm}
\label{tab:poolscaling}
\small
\setlength{\tabcolsep}{6pt}
\begin{tabular}{crccl}
\toprule
pairs & one-step val.\ loss & band exit $\uparrow$ & strict $\uparrow$ & per condition \\
\midrule
\phantom{0}60 & $9.7\times 10^{-5}$ & 388 & 368 & [278, 368, 371] \\
120 & $1.05\times 10^{-4}$ & 317 & 236 & [219, 236, 448] \\
242 & $1.21\times 10^{-4}$ & 650 & 319 & [140, 319, 431] \\
296 & $1.21\times 10^{-4}$ & \textbf{661} & \textbf{569} & [558, 569, 613] \\
\bottomrule
\end{tabular}
\end{table}

%% file: tables/table2b_oneshot_family.tex
\begin{table}[!htb]
\centering
\caption{One-shot family. Every configuration is fine-tuned from the baseline model on a reduced pool (rollouts to failure, three initial conditions, medians). The full-pool row is the seed whose band exit equals the five-seed median.}
\vspace{0.15cm}
\label{tab:oneshot}
\footnotesize
\setlength{\tabcolsep}{5pt}
\begin{tabular}{ccccccc}
\toprule
on-policy & surrogate-attack & DNS-probe & band exit & freeze & strict & failure mode \\
\midrule
\checkmark & \checkmark & \checkmark & 524 & 173 & 173 & freeze \\
\checkmark & -- & \checkmark & 639 & 636 & 636 & freeze (late) \\
\checkmark\ ($-$depths 16, 32) & \checkmark & \checkmark & 534 & 294 & 294 & freeze \\
-- & \checkmark & \checkmark & 54 & --- & 54 & blow-up \\
\bottomrule
\end{tabular}
\end{table}

%% file: tables/table2_component_ladder.tex
\begin{table}[!htb]
\centering
\caption{Component ablation, all eight combinations of the three generators, plus the main pool extended to depths 64 and 96 in row 5 (rollouts to failure, three initial conditions, medians). Row 1 is the single-seed baseline run, the bold row and row 5 are five-seed medians (Table~\ref{tab:seeds}), and the rest are single runs. The \emph{base} column gives the model each row was fine-tuned from, since the pool is assembled in stages and the leave-one-out rows below the rule therefore do not share one: rows 2 to 4 are those stages in order, the row without DNS-probe states is row 3 itself, and the two remaining leave-one-out rows retrain the last stage on the reduced pool. Table~\ref{tab:oneshot} repeats the comparison from a single fixed base.}
\vspace{0.15cm}
\label{tab:ladder}
\scriptsize
\setlength{\tabcolsep}{5pt}
\begin{tabular}{cccccccc}
\toprule
on-policy & surrogate-attack & DNS-probe & base & band exit & freeze & strict & failure mode \\
\midrule
-- & -- & -- & one-step & 13 & --- & 13 & blow-up \\
\checkmark & -- & -- & row 1 & 674 & 86 & 86 & freeze \\
\checkmark & \checkmark & -- & row 2 & 377 & 139 & 139 & freeze \\
\checkmark & \checkmark & \checkmark & row 3 & \textbf{721} & --- & \textbf{721} & band exit (late) \\
\checkmark\ ($+$depths 64, 96) & \checkmark & \checkmark & row 3 & 796 & 635 & 635 & freeze (late) \\
\midrule
-- & \checkmark & -- & row 1 & 76 & --- & 76 & blow-up \\
-- & -- & \checkmark & row 1 & 18 & --- & 18 & blow-up \\
\checkmark & -- & \checkmark & row 2 & 396 & --- & 396 & band exit (late) \\
-- & \checkmark & \checkmark & row 2 & 54 & --- & 54 & blow-up \\
\bottomrule
\end{tabular}
\end{table}

%% file: tables/table6_attack_grid.tex
\begin{table}[!htb]
\centering
\caption{Attack objectives ($2^3$ grid, 66 pairs per cell) and the solver-in-the-loop control. Band-exit step over a 200-step window. $^{\dagger}$single initial condition.}
\vspace{0.15cm}
\label{tab:attacks}
\small
\begin{tabular}{lcc}
\toprule
\textbf{objectives} & band exit (3 ICs) & median \\
\midrule
freeze-inducing & [9, 11, 10] & 10 \\
amplification-inducing & [51, 53, 57] & 53 \\
physics-violating & [50, 53, 58] & 53 \\
freeze + amplification$^{\dagger}$ & [63] & 63 \\
freeze + violation & [52, 53, 63] & 53 \\
amplification + violation & [59, 61, 63] & 61 \\
all three & [60, 76, 86] & \textbf{76} \\
\midrule
error maximization (solver in loop) & [45, 45, 70] & 45 \\
\bottomrule
\end{tabular}
\end{table}

%% file: tables/table3_depth_ablation.tex
\begin{table}[!htb]
\centering
\caption{On-policy depth ablation. Top: matched budget, only the maximum rollout depth of the on-policy states varies. Bottom: full budget. Single runs, except the two main-model rows (five-seed medians, Table~\ref{tab:seeds}).}
\vspace{0.15cm}
\label{tab:depth}
\small
\begin{tabular}{lccc}
\toprule
\textbf{max.\ on-policy rollout depth} & band exit & strict & freeze events (of 3) \\
\midrule
$\leq 8$ & 180 & 180 & 0 \\
$\leq 16$ & 406 & 330 & 1 \\
$\leq 32$ & 384 & \textbf{355} & 2 \\
$\leq 64$ & 573 & 308 & 3 \\
$\leq 96$ (all depths) & 438 & 180 & 3 \\
16 and 32 only & 421 & 358 & 1 \\
fine-tuning base (no on-policy pairs added) & 377 & 139 & 3 \\
\midrule
\textbf{main model (depths to 32)} & \textbf{721} & \textbf{721} & 0 \\
main model $+$ depths 64, 96 & 796 & 635 & 3 \\
\midrule
one-shot, full pool & 524 & 173 & 3 \\
one-shot $-$ depths 16, 32 & 534 & 294 & 3 \\
\bottomrule
\end{tabular}
\end{table}

%% file: tables/table7_recipe_knobs.tex
\begin{table}[!htb]
\centering
\caption{Training hyperparameters of the staged main model (721 on both criteria), one varied at a time (rollouts to failure, three initial conditions, medians). Single runs.}
\vspace{0.15cm}
\label{tab:knobs}
\small
\begin{tabular}{lcccc}
\toprule
\textbf{hyperparameter} & default & variant & band exit & strict \\
\midrule
learning rate & $10^{-4}$ & $5\times10^{-5}$ & 436 & 379 \\
pair-batch ratio & 0.3 & 0.2 & 613 & 218 \\
fine-tuning epochs & 24 & 4 & 465 & 86 \\
\bottomrule
\end{tabular}
\end{table}

%% file: tables/table4_label_paradigm_cross.tex
\begin{table}[!htb]
\centering
\caption{Label control (rollouts to failure, three initial conditions, medians). $^{\dagger}$same baseline model and fine-tuning schedule, only the training pairs differ. $^{\ddagger}$matched DNS budget, on the staged schedule of row 1. Row 1 is the five-seed median of Table~\ref{tab:seeds}. The rest are single runs.}
\vspace{0.15cm}
\label{tab:labels}
\footnotesize
\setlength{\tabcolsep}{4pt}
\begin{tabular}{lllcc}
\toprule
\textbf{states} & \textbf{labels} & \textbf{loss protocol} & band exit & strict \\
\midrule
off-attractor (OAS staged) & own future $S(\tilde{x})$ & 4-step unrolled on pairs & \textbf{721} & \textbf{721} \\
off-attractor (OAS one-shot)$^{\dagger}$ & own future $S(\tilde{x})$ & 4-step unrolled on pairs & 524 & 173 \\
clean windows, fine-tuning schedule$^{\dagger}$ & clean future $S(x)$ & 4-step unrolled & 38 & 38 \\
clean windows, $984$ new DNS frames$^{\ddagger}$ & clean future $S(x)$ & 4-step unrolled on pairs & 47 & 47 \\
clean windows, unrolled schedule & clean future $S(x)$ & 4-step unrolled & 259 & 259 \\
unrolled base $+$ attack pairs & mixed & 4-step unrolled & 99 & 99 \\
\bottomrule
\end{tabular}
\end{table}

%% file: tables/table8_lambda_probes.tex
\begin{table}[!htb]
\centering
\caption{One-step amplification $\lambda$ of the dynamics along four directions, measured with the DNS at relative amplitude $10^{-3}$.}
\vspace{0.15cm}
\label{tab:probes}
\small
\begin{tabular}{lc}
\toprule
\textbf{direction} & $\lambda$ \\
\midrule
worst (power iteration) & 1.036 \\
random isotropic & 0.417 \\
model's actual error direction & 0.94 \\
spectrum-shaped & 1.016 \\
\bottomrule
\end{tabular}
\end{table}

%% file: tables/table12_lambda_configs.tex
\begin{table}[!htb]
\centering
\caption{One-step amplification of the learned map along its worst direction and along a random one, with the failure times of the same run, for the seven configurations of Fig.~\ref{fig:mechanism}c. Each is the single run whose failure times the table reports. Both directions are probed at relative amplitude $10^{-3}$, as in Table~\ref{tab:probes}. The random-direction column is not bolded: it is flat across configurations and far from the $0.417$ of the dynamics, so it distinguishes nothing (Section~\ref{sec:mechanism}).}
\vspace{0.15cm}
\label{tab:lambdaconfigs}
\small
\begin{tabular}{lcccc}
\toprule
\textbf{configuration} & worst dir. & random dir. & band exit & strict \\
\midrule
DNS-probe states only & 3.17 & 1.03 & 18 & 18 \\
Baseline U-Net & 2.86 & 1.04 & 13 & 13 \\
Noise injection & 2.61 & 1.03 & 29 & 29 \\
Pushforward trick & 2.28 & 1.03 & 24 & 24 \\
Surrogate-attack states only & 2.10 & 1.03 & 76 & 76 \\
Unrolled training & 1.16 & 1.03 & 259 & 259 \\
\textbf{OAS (ours)} & \textbf{1.01} & 1.01 & \textbf{721} & \textbf{721} \\
\bottomrule
\end{tabular}
\end{table}

%% file: tables/table5_backbones.tex
\begin{table}[!htb]
\centering
\caption{Clean-trained backbones. Rollouts to failure from three initial conditions, medians with the per-condition values in brackets. Single runs.}
\vspace{0.15cm}
\label{tab:backbones}
\footnotesize
\setlength{\tabcolsep}{5pt}
\begin{tabular}{lcccc}
\toprule
\textbf{backbone} & nRMSE@1 & nRMSE@15 & band exit & strict \\
\midrule
U-Net (baseline)                    & 0.049 & 3.498 & \phantom{0}13 [\phantom{0}12, \phantom{0}13, \phantom{0}15] & 13 \\
FNO \citep{li2021fno}               & 0.102 & 2.120 & 408 [352, 408, 466] & 17 [11, 17, 18] \\
TFNO \citep{kossaifi2024mgtfno}     & 0.128 & 2.941 & \phantom{0}33 [\phantom{0}31, \phantom{0}33, \phantom{0}34] & 19 [18, 19, 23] \\
DPOT \citep{hao2024dpot}            & 0.164 & 5.831 & \phantom{0}11 [\phantom{00}7, \phantom{0}11, \phantom{0}12] & 11 \\
FactFormer \citep{li2023factformer} & 0.119 & 5.274 & \phantom{00}6 [\phantom{00}5, \phantom{00}6, \phantom{00}7] & \phantom{0}6 \\
\bottomrule
\end{tabular}
\end{table}

%% file: tables/table13_oas_backbones.tex
\begin{table}[!htb]
\centering
\caption{Off-attractor supervision on the two candidate backbones, at rollout step 15 and on the strict criterion, over rollouts to failure from three initial conditions. The clean-trained rows are the single run of Table~\ref{tab:backbones} (FNO) and the three-seed median of Table~\ref{tab:main} (U-Net). Fine-tuned rows are seed medians. Brackets show individual seed results or their range.}
\vspace{0.15cm}
\label{tab:oasbackbones}
\small
\setlength{\tabcolsep}{6pt}
\begin{tabular}{llcc}
\toprule
\textbf{backbone} & & nRMSE@15 $\downarrow$ & strict $\uparrow$ \\
\midrule
FNO   & clean-trained     & 2.120 & \phantom{00}17 \\
      & $+$ OAS (3 seeds) & 1.141 [1.077, 1.141, 1.232] & \phantom{00}35 [34, 35, 37] \\
      & gain              & $1.9\times$ & $2.1\times$ \\
\midrule
U-Net & clean-trained     & 3.498 & \phantom{00}21 \\
      & $+$ OAS (5 seeds) & \textbf{0.740} & \textbf{721} [705--784] \\
      & gain              & $\mathbf{4.7\times}$ & $\mathbf{34\times}$ \\
\bottomrule
\end{tabular}
\end{table}

%% file: tables/table11_second_regime.tex
\begin{table}[!htb]
\centering
\caption{Second regime, $\nu=0.010$ ($\mathrm{Re}_\lambda\!\approx\!35$), with its own dataset and pool. Rollouts to failure from three initial conditions, medians with the per-condition values in brackets. Single runs.}
\vspace{0.15cm}
\label{tab:regime2}
\small
\begin{tabular}{lcccc}
\toprule
 & pairs & epochs & band exit $\uparrow$ & strict $\uparrow$ \\
\midrule
Baseline U-Net & 0 & 24 & 77 [59, 77, 84] & 77 \\
\textbf{OAS (one-shot)} & 132$^{\ast}$ & 24 & \textbf{583} [503, 583, 589] & \textbf{540} [485, 540, 545] \\
\midrule
gain over baseline & & & $7.6\times$ & $7.0\times$ \\
\bottomrule
\multicolumn{5}{l}{\footnotesize $^{\ast}$54 on-policy, 54 surrogate-attack, 24 DNS-probe pairs.}
\end{tabular}
\end{table}

%% file: tables/table10_headline_robustness.tex
\begin{table}[!htb]
\centering
\caption{Seed robustness. Rollouts continued to failure, from three initial conditions. Median over seeds, with the per-seed medians in brackets.}
\vspace{0.15cm}
\label{tab:seeds}
\small
\begin{tabular}{lccc}
\toprule
 & seeds & band exit $\uparrow$ & strict $\uparrow$ \\
\midrule
Baseline U-Net & 3 & 21 [13, 22, 21] & 21 \\
Unrolled training & 3 & 89 [259, 52, 89] & 89 [259, 52, 89] \\
OAS one-shot & 5 & 524 [524, 399, 181, 774, 617] & 247 [173, 399, 181, 247, 617] \\
\textbf{OAS staged (main)} & 5 & \textbf{721} [705, 719, 721, 723, 784] & \textbf{721} [705, 719, 721, 723, 784] \\
OAS staged $+$ depths 64, 96 & 5 & 796 [685, 794, 796, 820, 833] & 635 [583, 593, 635, 647, 718] \\
\bottomrule
\end{tabular}
\end{table}